\documentclass[runningheads]{llncs}

\usepackage{eccv}

\usepackage{eccvabbrv}

\usepackage{graphicx}
\usepackage{booktabs}

\usepackage[accsupp]{axessibility}  

\usepackage{amsmath,amsfonts,bm}

\def\eqref#1{equation~\ref{#1}}

\def\1{\bm{1}}

\def\vh{{\bm{h}}}

\def\vm{{\bm{m}}}
\def\vn{{\bm{n}}}

\def\vv{{\bm{v}}}
\def\vw{{\bm{w}}}
\def\vx{{\bm{x}}}
\def\vy{{\bm{y}}}
\def\vz{{\bm{z}}}
\def\vepsilon{{\bm{\epsilon}}}

\def\mI{{\bm{I}}}

\DeclareMathAlphabet{\mathsfit}{\encodingdefault}{\sfdefault}{m}{sl}
\SetMathAlphabet{\mathsfit}{bold}{\encodingdefault}{\sfdefault}{bx}{n}

\def\maskencoder{{\mathcal{P}_\phi}}
\def\eqref#1{Eq.~(\ref{#1})}

\usepackage[utf8]{inputenc} 
\usepackage[T1]{fontenc}    
\usepackage{url}            
\usepackage{booktabs}       
\usepackage{amsfonts}       
\usepackage{nicefrac}       
\usepackage{microtype}      
\usepackage{xcolor}         
\usepackage{float}

\usepackage{lipsum}
\usepackage{graphicx}
\usepackage{wrapfig}
\usepackage{algorithm}
\usepackage{algpseudocode}
\usepackage{thmtools,thm-restate}
\usepackage{colortbl}
\usepackage{makecell}
\usepackage{tcolorbox}

\def\eqref#1{Eq.~(\ref{#1})}

\usepackage{animate}
\usepackage{tabularx}

\usepackage{color}
\usepackage{colortbl}
\definecolor{tabfirst}{rgb}{1, 0.7, 0.7} 
\definecolor{tabsecond}{rgb}{1, 0.85, 0.7} 
\definecolor{tabthird}{rgb}{1, 1, 0.7} 

\usepackage{caption}
\usepackage{subcaption}
\usepackage{graphicx}
\usepackage{booktabs} 
\usepackage{adjustbox} 

\usepackage{hyperref}

\usepackage{orcidlink}

\begin{document}

\title{InverseCrafter: Efficient Video ReCapture as a Latent Domain Inverse Problem}

\titlerunning{InverseCrafter}
\author{Yeobin Hong$^*$\inst{1}\orcidlink{0009-0001-2060-2207} \and
Suhyeon Lee$^*$\inst{1}\orcidlink{0009-0002-1476-6480} \and
Hyungjin Chung$^\dagger$\inst{2,3}\orcidlink{0000-0003-3202-0893} \and
Jong Chul Ye$^\dagger$\inst{1}\orcidlink{0000-0001-9763-9609}}

\authorrunning{Hong $\&$ Lee et al.}

\institute{
Graduate School of AI, KAIST, South Korea
\and
EverEx, South Korea \\
\and
Korea University, South Korea \\
$*$ Equal contribution, $\dagger$ Co-correspondence}

\maketitle
\begin{figure}
    \centering
    \vspace{-0.7cm}
    \captionsetup{type=figure}
    \includegraphics[width=\textwidth]{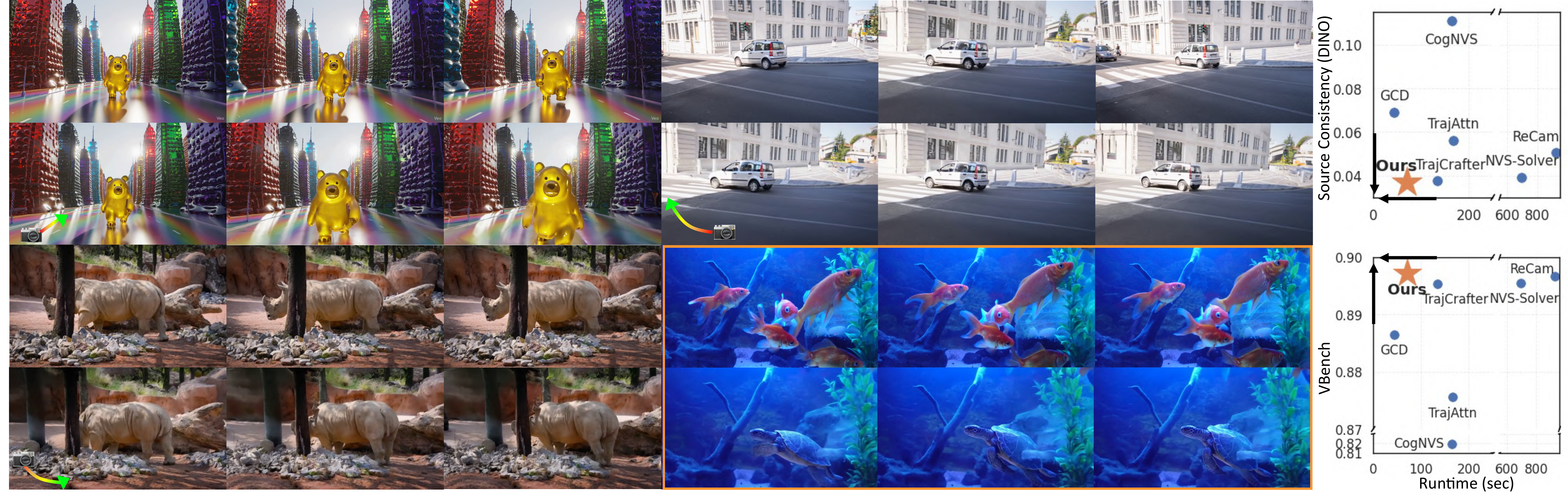}
    \vspace{-0.8cm}
    \captionof{figure}{\textbf{Representative video} on camera control \textit{("zoom in", "arc left", "arc right")}. Inpainting with editing \textit{("goldfish" to "turtle") in the right bottom box}.}
    \label{fig:represent}
    \vspace{-1.0cm}
\end{figure}%
\begin{abstract}
Recent approaches in controllable novel view video generation often rely on fine-tuning pre-trained Video Diffusion Models (VDMs). This dominant paradigm is computationally expensive and frequently suffers from catastrophic forgetting of the model's original generative priors. To address this challenge,  here we propose \textbf{InverseCrafter}, a VDM training-free framework that reformulates novel view video generation as an inpainting-based inverse problem in the latent space, eliminating the need for any annotated 4D training data. The core of our method is to establish operator equivalence by employing a lightweight latent mask encoder to define a latent-domain masking operation via a continuous, multi-channel representation. This principled representation faithfully models the forward process in the latent domain, enabling efficient, backpropagation-free solvers while bypassing the costly bottleneck of repeated VAE operations. InverseCrafter achieves high-fidelity, spatio-temporally coherent novel view synthesis with near-zero additional inference overhead and excels at general-purpose video inpainting and editing by fully preserving the pre-trained VDM's generative capabilities.
\end{abstract}
\section{Introduction}
Generating controllable novel view videos is a long-standing goal in computer vision, with applications in AR/VR, filmmaking, and digital twins. Recent Video Diffusion Models (VDMs) \cite{ho2022video, blattmann2023stable, videoworldsimulators2024, kong2024hunyuanvideo, wan2025wanopenadvancedlargescale} have achieved remarkable progress in generating photorealistic and diverse videos.
However, their control mechanism is typically limited to coarse inputs such as text descriptions or a conditioning first frame, offering minimal control over camera motion or scene geometry. This limitation prevents their direct application to novel view synthesis, which requires consistent rendering under user-specified camera trajectories.

One branch of research that has emerged to address the challenge of novel view synthesis is the framing of the problem as target-view recapture given an input video: given a input video, synthesize novel views over time while following a target camera trajectory (\eg, camera control).
For example, a line of work fine-tunes VDMs to condition on prospective cameras~\cite{van2024generative, bai2025recammastercameracontrolledgenerativerendering, wu2024cat4d}, which requires large-scale 4D datasets and still generalizes poorly to unseen camera trajectories.
Another branch~\cite{ren2025gen3c3dinformedworldconsistentvideo, gu2025diffusionshader3dawarevideo, xiao2024trajectory, yu2025trajectorycrafterredirectingcameratrajectory} reframes the task as 3D-aware inpainting problem and solve it via VDM fine-tuing.  Specifically, these methods leverage 3D reconstruction priors (\eg, depth estimation or point tracking) to create large-scale, pseudo-annotated datasets, and then fine-tune the VDM to inpaint the occluded regions. Finetuning, however, often creates a strong dependency on the specific 3D reconstruction model used during training. Moreover, finetuning often leads to overfitting and catastrophic forgetting, degrading the pretrained model’s original text-to-video generation capability and limiting its usefulness as a general generative prior. Furthermore, some methods require further augmenting the VDM architecture (\eg, adding new attention blocks~\cite{xiao2024trajectory, yu2025trajectorycrafterredirectingcameratrajectory}), substantially increasing inference latency. Although, per-video optimization approaches~\cite{jeong2025reangleavideo4dvideogeneration} avoid large-scale retraining, they overfit to individual scenes and fail to generalize.

\begin{figure*}[t]
    \centering
    \includegraphics[width=\linewidth]{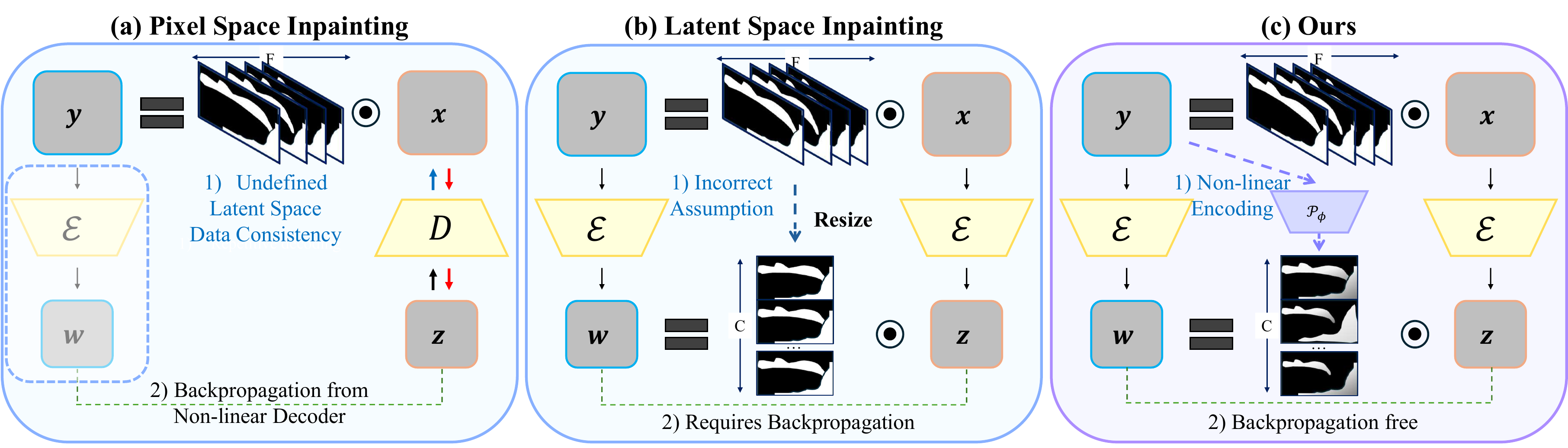}
    \caption{
    (a) Pixel-space inpainting~\cite{song2024solving,rout2024solving,kim2025flowdps} requires per-timestep VAE decoding, increasing computational cost. (b) Existing latent space inpainting methods~\cite{yu2025trajectorycrafterredirectingcameratrajectory, jeong2025reangleavideo4dvideogeneration, you2024nvs} downsamples the pixel space mask via spatio-temporal interpolation. This process results in a single-channel, binary mask that is uniformly broadcast across all $C$ latent channels, ignoring their distinct feature representations and leading to information loss. These methods typically rely on computationally heavy backpropagation. (c) InverseCrafter computes a continuous, $C$-channel latent mask, enabling efficient latent-space guidance.
    }     
    \vspace{-0.3cm}
    \label{fig:concept}
\end{figure*}

A promising training-free alternative is to formulate novel view video generation as an inpainting-based inverse problem at inference time, where the warped video provides partial measurements and a pretrained VDM inpaints the unknown pixels at the inference time. However, adapting diffusion inverse solvers from the image domain to video inpainting is challenging because modern diffusion models operate in a compressed spatio-temporal latent space. First, naively performing pixel-space inpainting (Fig.~\ref{fig:concept}(a))~\cite{song2024solving,rout2024solving,kim2025flowdps} requires per-timestep decoding, incurring significant computational overhead and instability due to the nonlinear decoder~\cite{raphaeli2025silo}.

To mitigate this issue, recent video inpainting methods -- including training-based methods~\cite{jeong2025reangleavideo4dvideogeneration,yu2025trajectorycrafterredirectingcameratrajectory} and training-free inverse solvers~\cite{you2024nvs, park2025zero4dtrainingfree4dvideo} (Fig.~\ref{fig:concept}(b)), approximate the pixel-space inpainting problem in the latent space by naively resizing the pixel-space mask to the latent domain. This removes the need for VAE decoding and results in a linearized inverse problem in the full latent space. Although computationally efficient in practice, this approach assumes that spatio-temporal structures and relationships are faithfully preserved in the compressed latent representation -- an assumption that often fails, resulting in degraded quality and visible artifacts. Furthermore, despite the linearized formulation, existing video latent inverse solvers do not fully exploit general-purpose linear solvers~\cite{park2025zero4dtrainingfree4dvideo}; instead, they rely on architecture-specific algorithms tailored to particular VDMs~\cite{yesiltepe2025dynamicviewsynthesisinverse} or even resort to computationally expensive VDM backpropagation~\cite{you2024nvs}.

To address this gap, we present \textbf{InverseCrafter}, a framework that formulates controllable novel view video generation as a  latent-domain continuous inpainting-based inverse problem at inference time under a pretrained VDM prior.
InverseCrafter requires no VDM fine-tuning and no additional 4D datasets, and it incurs negligible inference overhead, yet it enables controllable novel view video generation from a single input video under arbitrary camera trajectories.

Specifically, consider the pixel-space forward process $\vy = \vx \odot \vm$.
Here, $\vy$ and $\vx$ denote the measurement and the unknown novel view video, respectively, after the camera view change, and $\vm$ denotes  the mask indicating the unconvered regions induced by the view change.
The key idea of InverseCrafter is to establish \textit{operator equivalence} between the pixel-space measurement process and its latent-space counterpart, as highlighted in~\cite{raphaeli2025silo}. More specifically, rather than of naively resizing the pixel-space mask, we learn a latent-space equivalent masking operator that faithfully reproduces the pixel-space forward process $\vy = \vx \odot \vm$ after VAE encoding. To this end, we train a lightweight \textit{latent mask encoder} $\maskencoder$ to predict a continuous $C$-channel latent mask $\vh$. This learned masking operator enables the latent operation $\vw = \vz \odot \vh$ to be operator-equivalent to the original pixel-space masking process. This formulation enables efficient latent-space inverse solvers, allowing guidance to be performed entirely in the VDM latent space without per-step VAE decoding or costly backpropagation, significantly reducing computation while maintaining high synthesis quality through more accurate operator modeling.

Our main contributions are summarized as follows:
\begin{itemize}
    \item We reformulate controllable  video generation as an inverse problem at inference time, leveraging pretrained VDM priors without task-specific fine-tuning or additional 4D datasets.
    \item We introduce a mask encoder to establish operator equivalence via a continuous mask in the latent space, which enables leveraging efficient, backpropagation-free inverse problem solvers without any decoding into pixel space.
    \item Our framework achieves high-fidelity, spatio-temporally coherent  synthesis with near-zero additional inference overhead over standard diffusion sampling.
\end{itemize}
%
\section{Preliminaries}
\subsection{Video Inpainting for  Generation}
Given a partially observed target-view video obtained through geometric warping, the occluded regions must be filled in a temporally coherent manner. When latent diffusion models (LDMs) are used as priors, this completion is performed in the compressed latent space~\cite{yu2025trajectorycrafterredirectingcameratrajectory, you2024nvs, wang2023zeroshot, jeong2025reangleavideo4dvideogeneration}. 
For instance, NVS-Solver \cite{you2024nvs} employ diffusion Posterior Sampling (DPS) in the latent space at inference time. 
Zero4D \cite{park2025zero4dtrainingfree4dvideo} formulates the 4D video generation as a video interpolation problem in the latent domain by employing mask-based interpolation between the generated output and the measurement.

A critical limitation of these approaches is their reliance on models like Stable Video Diffusion (SVD) \cite{blattmann2023stable}, whose VAE performs spatial compression only. Both project the degradation operator (pixel space mask) into the latent space via {\em simple average pooling} over the spatial dimensions.
The challenge of correctly formulating the measurement and masking operation in the latent space is a common, unaddressed issue. 
This problem of binary resizing is significantly exacerbated when it comes to modern Video Diffusion Models (VDMs). Since modern VDMs utilize 3D VAEs for spatio-temporal compression, mapping a 2D pixel space mask to a 3D latent space is ill-defined.

\subsection{Diffusion Models for Inverse Problems}

Consider the following forward model:
\begin{equation}\label{eq:inverse}
    \vy = \mathcal{A}(\vx) + \vn,
\end{equation}
where $\mathcal{A}$ denotes the forward operator and $\vn$ is additive noise. The goal of the inverse problem is to recover the unknown signal $\vx$ from the corrupted measurement $\vy$. Although diffusion models are trained to sample from the prior distribution $p(\vx)$, diffusion-based inverse solvers modify the reverse diffusion dynamics at inference time to approximate posterior sampling, \textit{i.e.}, $\vx \sim p(\vx \mid \vy)$~\cite{chung2025diffusion}. Most zero-shot approaches guide the reverse diffusion process to reduce the following data consistency term $\|\vy - \mathcal{A}(\hat{\vx}_{0|t})\|^2$, where $\hat{\vx}_{0|t}=\mathbb{E}[\vx_0|\vx_t]$ denotes the denoised estimate at timestep $t$.

Unfortunately, applying such solvers to LDMs—which dominate modern generative modeling—requires additional considerations. Existing latent-space solvers~\cite{rout2024solving,chung2024prompt,kim2025flowdps,spagnoletti2025latino} typically decode the intermediate latent variable $\hat{\vz}_{0|t}=\mathbb{E}[\vz_0|\vz_t]$ into the pixel space by decoding with the VAE decoder $\mathcal{D}$, and reduce the measurement consistency term in the decoded pixel space, $\|\vy - \mathcal{A}(\mathcal{D}(\hat{\vz}_{0|t}))\|^2$, to update the latent variable $\vz$. However, enforcing measurement consistency through the nonlinear decoder $\mathcal{D}$ introduces approximation errors and substantial computational overhead, often leading to over-smoothed or blurry reconstructions.

Recently, SILO~\cite{raphaeli2025silo} addressed this limitation by training a latent forward operator $H_\theta$ that approximates the pixel-space forward operator $\mathcal{A}$ through an operator-equivalent mapping between the pixel and latent domains. Specifically, this formulation enables measurement consistency to be evaluated directly in the latent space as $\|\mathcal{E}(\vy) - H_\theta(\hat{\vz}_{0|t})\|^2$, where $\mathcal{E}$ refers to the encoder of VAE, thereby eliminating repeated decoding to the pixel domain. Nevertheless, since $H_\theta$ is nonlinear, it cannot leverage efficient linear inverse solvers, and backpropagation through the neural network may reduce numerical stability. Moreover, training $H_\theta$ requires paired samples $(\vy,\vx)$, which becomes impractical in our setting where large-scale 4D datasets are scarce.

\section{InverseCrafter}
\subsection{Problem Formulation}
To solve the inverse problem of novel view synthesis, we must formulate the measurement $\vy$ based on 3D geometric constraints. This requires projection of the source video into a target camera view.
First, we establish a 3D representation of the scene. We process the source video $\vx^{src}_{1:F}$ frame-by-frame using a pre-trained monocular depth estimation network~\cite{hu2024depthcraftergeneratingconsistentlong} to acquire a per-frame estimated depth map $\mathbf{\hat{D}}_{1:F}$.
Next, for each frame $i$, we unproject the 2D pixel coordinates (from the image $\vx^{src}_i$) into a 3D point cloud $\mathbf{P}_i$ using the estimated depth $\mathbf{\hat{D}}_i$ and an intrinsic camera matrix $\mathbf{K}$. This back-projection function is denoted $\Pi^{-1}$:
\begin{align}
    \mathbf{P}_i = \Pi^{-1}(\vx^{src}_i, \mathbf{\hat{D}}_i, \mathbf{K}).
\end{align}
Once the scene is represented in 3D, we can render it from a new perspective. The 3D point cloud $\mathbf{P}_i$ is transformed into the target camera's coordinate system using a relative transformation matrix $\mathbf{T}_i$. These transformed points are then projected back into a 2D image $\vy_i$ using the forward projection function $\Pi$:
\begin{align}\label{eq:warp}
    \vy_i = \Pi(\mathbf{T}_i \cdot \mathbf{P}_i, \mathbf{K}).
\end{align}
The warped video $\vy := \vy_{1:F}$ represents the pixel-space \textit{measurement}, which contains occlusions. Our goal is  then to recover the target-view video $\vx := \vx_{1:F}$ from occluded measurement obtained through 3D warping:
\begin{align}
\label{eq:ip_pixel_space}
    \vy = \vm \odot \vx,\quad \vx,\vy,\vm \in \mathbb{R}^{N}
\end{align}
where $\odot$ denotes the element-wise product and $N = FHW$ is the pixel space resolution, and $\vm$ refers to a mask that represents the visiblity induced by the 3D warping process. This defines the pixel-space \textit{forward operator}.

\begin{figure*}[t]
    \centering
    \includegraphics[width=\linewidth]{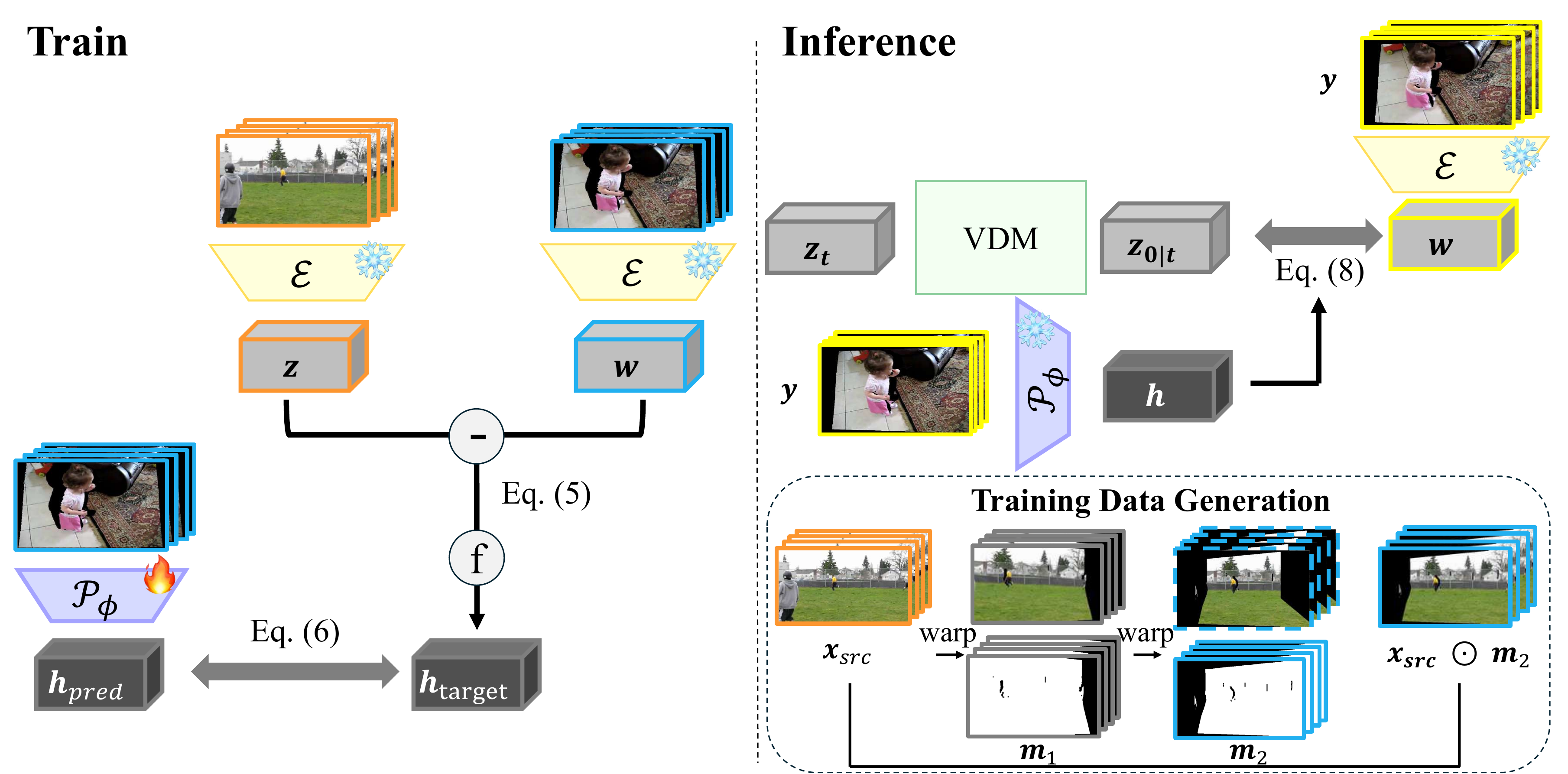}
    \vspace{-0.5cm}
    \caption{\textbf{Overview of InverseCrafter.} \textbf{(a)} $\maskencoder$ is trained to project the pixel space degradation operator to the latent domain. \textbf{(b)} During inference, $\vz_t$ is optimized at each $t$ to enforce data consistency \eqref{eq:latent_prox}, the latent mask derived from the $\maskencoder$.}
    \vspace{-0.5cm}
    \label{fig:method}
\end{figure*}
Recent VDMs~\cite{yang2024cogvideox, wan2025wanopenadvancedlargescale} introduce 3D VAEs that perform spatio-temporal compression, creating the need to reformulate the pixel-domain forward model in \eqref{eq:ip_pixel_space} in the latent domain to preserve operator equivalence between the pixel and latent domains, enabling efficient inverse problem solving in the latent space.
Image inpainting inverse solvers for LDMs~\cite{kim2024dreamsampler, kim2023regularization} typically construct latent masks by projecting pixel-space masks into the latent space via nearest-neighbor downsampling. This approximation is an ad-hoc adaptation for the LDM, originating from pixel-space diffusion inverse solvers. However, naive downsampling only loosely approximates the true degradation operator. The additional temporal compression in video models further exacerbates this mismatch, motivating a more principled treatment beyond these image-based approaches.

Nevertheless, recent video inpainting methods that operate directly in the latent domain~\cite{yu2025trajectorycrafterredirectingcameratrajectory, xiao2024trajectory, jeong2025reangleavideo4dvideogeneration, park2025zero4dtrainingfree4dvideo} largely adopt direct spatio-temporal projection (\ie, spatio-temporal interpolation) as a naive extension of image-based approaches. 
This practice has several limitations. 
First, it implicitly assumes that spatial structures in pixel space are preserved under the non-linear VAE encoder, which is generally false. Second, it treats the latent mask as uniform across channels, despite the fact that latent channels encode highly heterogeneous feature representations, as illustrated in Fig.~\ref{fig:concept}(c) (visualizations in Appendix~\ref{sec:mask_viz}). Third, some methods~\cite{jeong2025reangleavideo4dvideogeneration, yu2025trajectorycrafterredirectingcameratrajectory} adopt an overly conservative temporal rule: a latent region is marked as “known” only if its corresponding pixel region is visible in all frames within its temporal group. This strategy becomes a major bottleneck for videos with fast motion. If an object appears in only a subset of frames due to rapid movement or partial occlusion, the entire corresponding latent region is marked as “unknown”, which forces unnecessary inpainting and increases reliance on the generative model, as shown in Fig.~\ref{fig:ablation}(a).

\subsection{Generating Continuous Mask}
To address the mask projection inconsistencies outlined previously, we propose learning a lightweight neural network, dubbed the mask encoder $\maskencoder(\cdot): \mathbb{R}^N \rightarrow \mathbb{R}^{C \times M}$, where $C$ denotes the number of latent channels and $M = f h w$ represents the latent-space resolution. Then, our goal is to predict a latent mask that parameterizes a latent-domain masking operator approximating the operator-equivalent to the pixel-space masking. Specifically, the network takes as input the pixel measurement $\vy$ and outputs a continuous mask that can be applied {\em linearly} to the latent $\vz$, with the same Hadamard product operation as in \eqref{eq:ip_pixel_space}. 

Our training target is derived from the observation that autoencoders are effective at reconstructing degraded data \cite{raphaeli2025silo}. We define this ground-truth latent mask, $\vh \in [0, 1]^{C\times M}$, as the normalized difference between the latent encoding of the clean video, $\mathcal{E}(\vx)$, and the latent encoding of the masked video, $\mathcal{E}(\vm \odot \vx)$:
\begin{equation}
    \vh = 1 - \mathrm{f} \left( \mathcal{E}(\vx) - \mathcal{E}(\vm \odot \vx) \right),
\end{equation}
where $\mathrm{f}$ stands for normalization using an activation function, scaling the latent difference to a $[0, 1]$ range.

However, a primary challenge is obtaining training pairs, as for generating warped videos (\eqref{eq:warp}), the corresponding ``ground truth'' video $\vx$ is not available unlike in typical inverse problem settings. To resolve this, we employ a double-reprojection strategy from \cite{yu2025trajectorycrafterredirectingcameratrajectory} to synthesize the correctly masked counterpart $\vm \odot \vx$ from the clean video $\vx$, which provides the necessary pair to compute $\vh$ (See  Fig.~\ref{fig:method}-Right).

The mask encoder is then trained to predict this ground-truth latent mask directly from the pixel space measurement with a combination of the L1 loss and the Structural Similarity Index Measure (SSIM)~\cite{wang2004image} loss:
\begin{equation} \label{eq:mask_loss}
    \min_\phi \left( \left\| \maskencoder(\vy) - \vh \right\|_1 + \lambda \left( 1 - \text{SSIM}(\maskencoder(\vy), \vh) \right) \right)
\end{equation}
\begin{figure*}[!t]
\vspace{-0.5cm}
\centering
\begin{minipage}[t]{0.60\textwidth}
\begin{algorithm}[H]
\caption{InverseCrafter}
\label{alg:method}
\begin{algorithmic}[1]
\Require Source video $\vx^{src}$, 
pixel space measurement $\vy$,
pixel space mask $\vm$,
I2V flow-based model $\vv^\theta$, VAE encoder \& Decoder $\mathcal{E}/\mathcal{D}$, hyperparameter $\Gamma$, CG iteration $K$
\State $\vh \gets \maskencoder(\vy)$
\State \textbf{Initialization} $\vz_T \sim \mathcal{N}(0, \mathbf{I})$
\State $\vw \gets \mathcal{E}(\vy)$
\State $\mathbf{H} \gets \text{diag}(\vh)$        
\For{$t: 1\rightarrow 0$}
    \State $\triangleright$ Tweedie denoising
    \State $\hat{\vz}_{0|t} = \vz_t - \vv^\theta(\vz_t) t$
    \State $\hat{\vz}_{1|t} = \vz_t + \vv^\theta(\vz_t) (1-t)$
    \If{$t \in \Gamma $}
        \State $\triangleright$ Data consistency
        \State $\hat{\mathbf{z}}(\mathbf{y}) \leftarrow \text{CG}(\mathbf{I} + \gamma\mathbf{H}^\top\mathbf{H}, \hat{\vz}_{0|t} + \gamma \mathbf{H}^\top\vw, \hat{\mathbf{z}}_{0|t}, K)$
        \State $\vz_{t-1} \gets \text{ODESolve}(\hat{\vz}(\vy), \hat{\vz}_{1|t})$
    \Else
        \State $\vz_{t-1} \gets \text{ODESolve}(\hat{\vz}_{0|t}, \hat{\vz}_{1|t})$
    \EndIf
\EndFor
\State $\vx \gets \mathcal{D}(\vz_0)$
\end{algorithmic}
\end{algorithm}
\end{minipage}\hfill
\begin{minipage}[t]{0.35\textwidth}
\begin{algorithm}[H]
\caption{Conjugate Gradient (CG)}
\label{alg:cg}
\begin{algorithmic}[1]
\Require $\mathbf{A}, \mathbf{y}, \mathbf{x}_0, M$
\State $\mathbf{r}_0 \leftarrow \mathbf{b} - \mathbf{A}\mathbf{x}_0$
\State $\mathbf{p}_0 \leftarrow \mathbf{b}_0$
\For{$i = 0 : K - 1$ \textbf{do}}
    \State $\alpha_k \leftarrow \frac{\mathbf{r}_k^\top \mathbf{r}}{\mathbf{p}_k^\top \mathbf{A}\mathbf{p}_k}$
    \State $\mathbf{x}_{k+1} \leftarrow \mathbf{x}_k + \alpha_k \mathbf{p}_k$
    \State $\mathbf{r}_{k+1} \leftarrow \mathbf{r}_k - \alpha_k \mathbf{A}\mathbf{p}_k$
    \State $\beta_k \leftarrow \frac{\mathbf{r}_{k+1}^\top \mathbf{r}}{\mathbf{r}_k^\top \mathbf{r}_k}$
    \State $\mathbf{p}_{k+1} \leftarrow \mathbf{b}_{k+1} + \beta_k \mathbf{p}_k$
\EndFor
\State \textbf{return} $\mathbf{x}_K$
\end{algorithmic}
\end{algorithm}
\end{minipage}
\vspace{-0.5cm}
\end{figure*}
Unlike naive mask resizing, which generates a single-channel binary mask that is uniformly broadcast across all $C$ latent channels (ignoring their distinct feature representations), our mask encoder outputs a continuous, $C$-channel mask. This high-fidelity representation better captures the mask's influence across the latent features, leading to improved VAE reconstruction and superior measurement consistency in final generated outputs, as detailed in Tab.~\ref{fig:ablation_quantitative}.
\subsection{DDS Solver for Latent Inpainting with Continuous Masks}
To align with the latent diffusion model, we encode the pixel space measurement into the latent space using the VAE encoder $\mathcal{E}$. Then, under our noiseless linear setting of \eqref{eq:ip_pixel_space}, the inverse problem in the pixel domain transforms to an inverse problem in the latent space, i.e.
\begin{align}
\label{eq:ip_latent}
    \underbrace{\vw}_{\mathcal{E}(\vy)} = 
    \underbrace{\vh}_{\mathcal{P}_\phi(\vy)} \odot \underbrace{\vz}_{\mathcal{E}(\vx)},\quad \vz,\vw,\vh \in \mathbb{R}^{C \times M}.
\end{align}
Notice that the design of our mask encoder $\maskencoder$ is different from that of~\cite{raphaeli2025silo}. The operator encoder used in~\cite{raphaeli2025silo} acts {\em non-linearly} on the encoded latent $\vz$ to produce the latent measurement $\vw$, whereas our mask encoder produces a continuous linear mask $\vh$, which can be applied linearly with a Hadamard product. Thanks to such a design choice, we are free from selecting {\em any} diffusion model-based inverse problem solvers, even the ones that are specialized for solving linear inverse problems in the pixel space (See appendix Sec.~\ref{sec:solvers}). Among them, we propose to leverage DDS~\cite{chung2024decomposed}. Namely, the data consistency step of DDS solves the following proximal optimization problem :
\begin{equation}
\label{eq:latent_prox}
    \min_\vz \frac{\gamma}{2}\|\vw - \vh \odot \vz\|_2^2 + \frac{1}{2}\|\vz - \hat{\vz}_{0|t}\|_2^2,
\end{equation}
which is followed by the ODE solver.
Here,
$\hat{\vz}_{0|t} := \mathbb{E}[\vz_0|\vz_t]$ is the estimated posterior mean with the flow model $\vv^\theta$, and $\gamma$ is a hyper-parameter that controls the proximal regularization. Following \cite{chung2024decomposed}, we solve \eqref{eq:latent_prox} using conjugate gradient (CG) iterations (Alg.~\ref{alg:cg}). Specifically, let $\mathbf{H} := \text{diag}(\vh)$. Then, \eqref{eq:latent_prox} can be solved with $\text{CG}(\mathbf{I} + \gamma\mathbf{H}^\top\mathbf{H}, \hat{\vz}_{0|t} + \gamma \mathbf{H}^\top\vw, \hat{\vz}_{0|t}, K)$ with $K$ iterations. The overall algorithm of InverseCrafter is shown in Alg.~\ref{alg:method}.
%
%
\section{Experimental Results}
\subsection{Evaluation Details}
\begin{figure*}[t]
    \centering
    \includegraphics[width=0.8\linewidth]{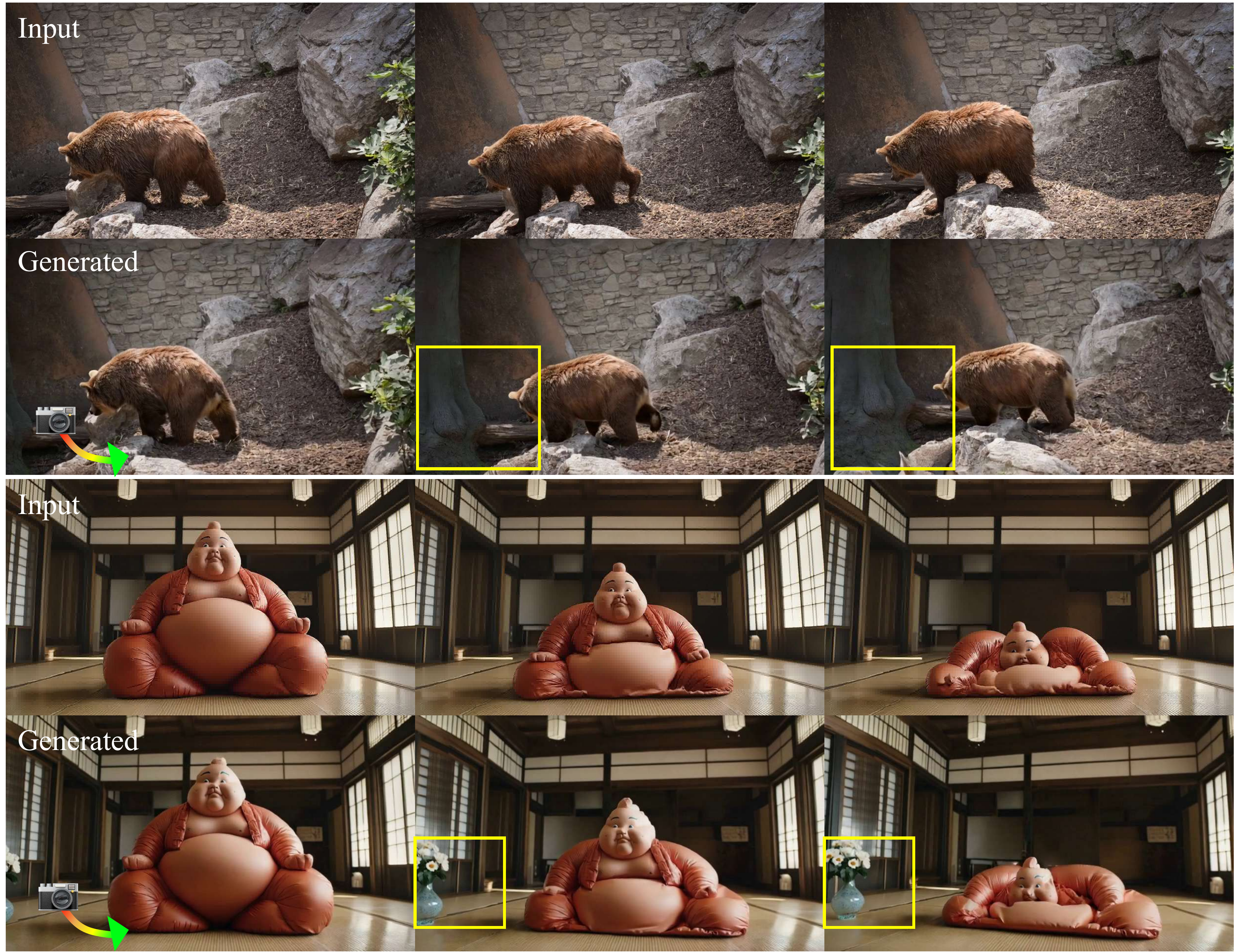}
    \caption{\textbf{Video camera control results with novel content generation.} (Up) "+a grey tree." (Bottom) "+a flower vase."}
    \label{fig:text_guidance}
\end{figure*}
\subsubsection{Task}
We evaluate our method on two major video inpainting tasks: video camera control and text-guided video inpainting. Our primary focus is the video camera control task, for which we use 1,000 videos and corresponding captions from the UltraVideo dataset~\cite{xue2025ultravideo}. We sample one of the six target camera trajectories for each video, where the target trajectories are selected following the work in \cite{jeong2025reangleavideo4dvideogeneration}. 
For text-guided video inpainting, we apply our proposed video inpainting method to replace objects in the video, guided by target text prompts and source object masks. This evaluation is conducted on 20 videos from the DAVIS~\cite{pont20172017} dataset, along with all available foreground masks. For each video, we generate five target editing prompts using GPT-5-mini~\cite{openai_gpt5mini_doc_2025}, while the corresponding source captions used for prompt construction are obtained using BLIP-2~\cite{li2023blip}.
\vspace{-0.5cm}
\subsubsection{Evaluation}
For video camera control task, we assess performance using a comprehensive set of metrics: measurement consistency - PSNR, LPIPS, SSIM, source consistency - DINO distance~\cite{caron2021emerging}, Fréchet Video Distance (FVD), and general video quality - subject consistency, temporal flickering, motion smoothness, and overall consistency from VBench~\cite{huang2024vbench++}.
The measurement consistency metric was compared by the warped video, frame by frame. DINO and FVD were calculated by reference to the source video datasets. For video inpainting tasks, we additionally evaluate the generated videos with the editing text-alignment metrics: VLM (see appendix \ref{sec:vllm} for details), CLIP Score~\cite{radford2021learning}, and Pick Score~\cite{kirstain2023pickapicopendatasetuser}.
\vspace{-0.5cm}

\subsection{Implementation Details}
\begin{table}[t]
\caption{Quantitative evaluation on the camera control task, showing source video consistency and generation quality together with runtime efficiency.}
\vspace{-0.3cm}
\label{tab:camera-control}
\centering
\setlength{\tabcolsep}{3pt}
\resizebox{\columnwidth}{!}{
\begin{tabular}{lccccccc}
\toprule
\textbf{Method} & \textbf{Runtime} & \multicolumn{3}{c}{\textbf{Measurement Consistency}} & \multicolumn{2}{c}{\textbf{Source Consistency}} & \textbf{VBench} \\
\cmidrule(r){2-2} \cmidrule(r){3-5} \cmidrule(r){6-7} \cmidrule(r){8-8}
& \small sec $\downarrow$ & \small PSNR $\uparrow$ & \small LPIPS $\downarrow$ & \small SSIM $\uparrow$ & \small DINO $\downarrow$ & \small FVD $\downarrow$ & Total Score $\uparrow$ \\
\cmidrule(r){1-1} \cmidrule(r){2-8}
ReCamMaster~\cite{bai2025recammastercameracontrolledgenerativerendering} & 926 & - & - & - & 0.0504 & 136.28 & 0.8967 \\
GCD~\cite{van2024generative} & \textbf{44} & - & - & - & 0.0691 & 250.85 & 0.8865 \\
TrajAttention~\cite{xiao2024trajectory} & 167 & 19.44 & 0.1405 & 0.6909 & 0.0562 & 233.40 & 0.8756 \\
TrajCrafter~\cite{yu2025trajectorycrafterredirectingcameratrajectory} & 134 & \underline{28.37} & \underline{0.0573} & \textbf{0.8942} & \underline{0.0376} & 120.03 & 0.8954 \\
NVS-Solver~\cite{you2024nvs} & 696 & 26.68 & 0.0816 & 0.8314 & 0.0393 & \textbf{96.90} & \underline{0.8955} \\
CogNVS~\cite{chen2026reconstructinpainttesttimefinetune} & 164 & 15.41 & 0.4339 & 0.4016 & 0.1108 & 2175.93 & 0.8193 \\
\cmidrule(r){1-1} \cmidrule(r){2-8}
\textbf{Ours} & \underline{71} & \textbf{29.35}	& \textbf{0.0485} & \underline{0.8827} &	\textbf{0.0359}	& \underline{99.73} & \textbf{0.8977} \\
\bottomrule
\end{tabular}
}
\end{table}
\begin{figure}[!ht]
    \centering
    \includegraphics[width=\linewidth]{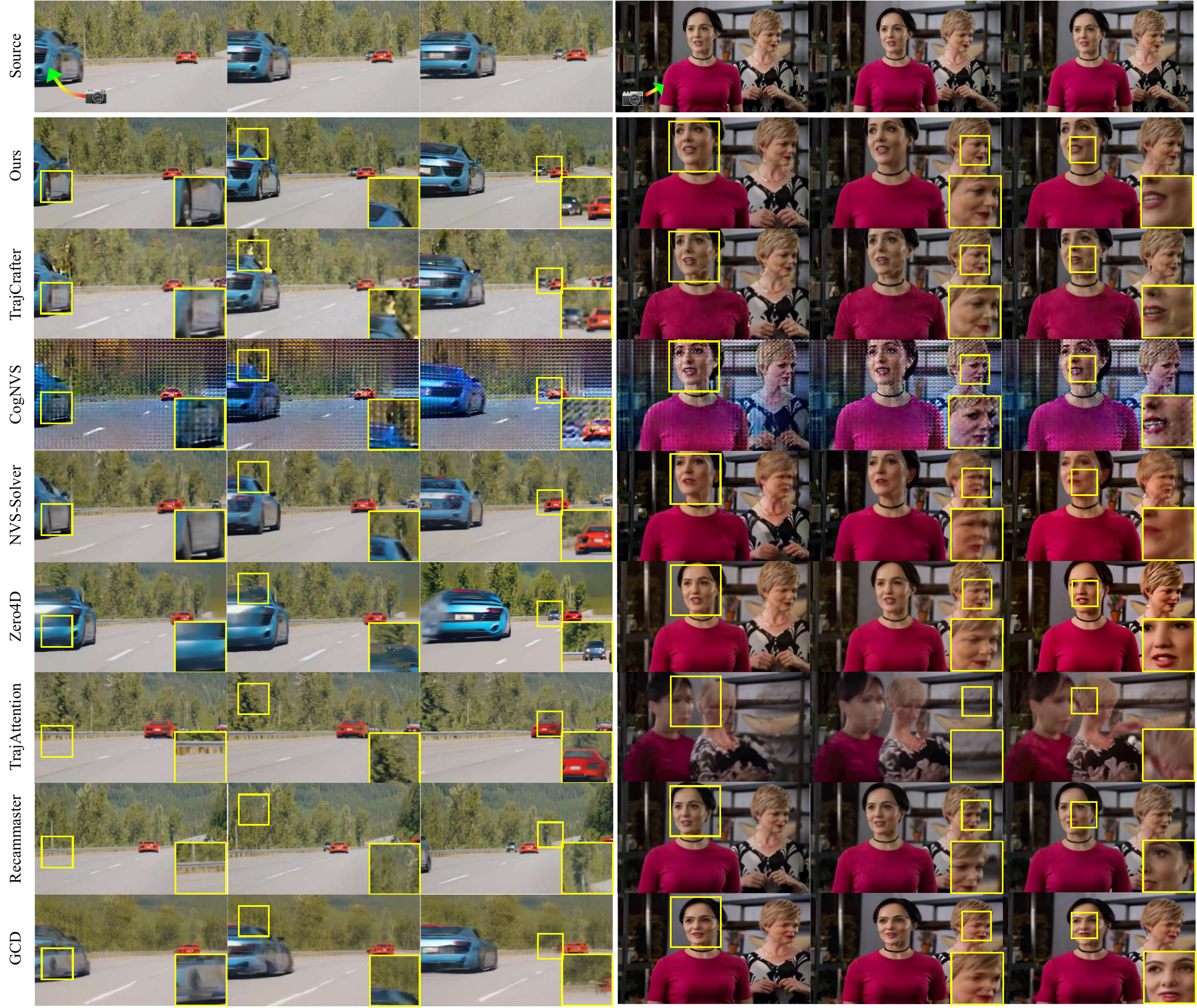}
    \vspace{-0.3cm}
    \caption{\textbf{Qualitative comparison of video camera control.} Camera trajectories are \textit{("arc left" "zoom in")}. Our method demonstrates a clear advantage in source consistency and semantically aligned generation. Insets provide magnified views of the regions marked by yellow boxes.}
    \label{fig:qualitative_reangle}
\end{figure}
\subsubsection{Training}
We train our model on the VidSTG \cite{zhang2020doesexistspatiotemporalvideo} dataset. For each of the 7,750 videos in the dataset, we generate 6 double-reprojection datasets from the selected camera trajectories, resulting in a total of 46,500 training samples. We utilize DepthCrafter~\cite{hu2024depthcraftergeneratingconsistentlong} for monocular depth estimation. Our model architecture is based on the Wan2.1 VAE, which we modify to create a lightweight model by reducing the channel dimension from 96 to 16, resulting in 1.5M parameters. The model is trained at a fixed resolution of 240$\times$416. We employ the AdamW optimizer~\cite{loshchilov2017decoupled} with a learning rate of $1 \times 10^{-4}$ and a weight decay of $3 \times 10^{-2}$. Training is conducted with a total batch size of 16, distributed as 4 samples per GPU across 4 GPUs, completed in 1 day.
\vspace{-0.5cm}
\subsubsection{Inference}
We use Wan2.1-Fun-V1.1-1.3B-InP~\cite{wan2025wanopenadvancedlargescale} at a resolution of 480$\times$832 for the main experiments.
For encoding masked videos during training or inference, as in \cite{jeong2025reangleavideo4dvideogeneration}, we infill the masked region to minimize the train-inference following gap \cite{saxena2023surprisingeffectivenessdiffusionmodels}. 
We use \cite{hu2024depthcraftergeneratingconsistentlong} for all training-free methods and TrajectoryCrafter for fair comparison to generate a depth map for each frame.
\vspace{-0.5cm}
\subsubsection{Compute Resources}
All experiments are conducted using NVIDIA RTX A6000 GPUs (48GB VRAM). A detailed runtime analysis is provided in Tab.~\ref{tab:camera-control}.
\subsection{Results}
\subsubsection{Video Camera Control}
\begin{figure}[!pt]
    \centering
    \includegraphics[width=0.8\linewidth]{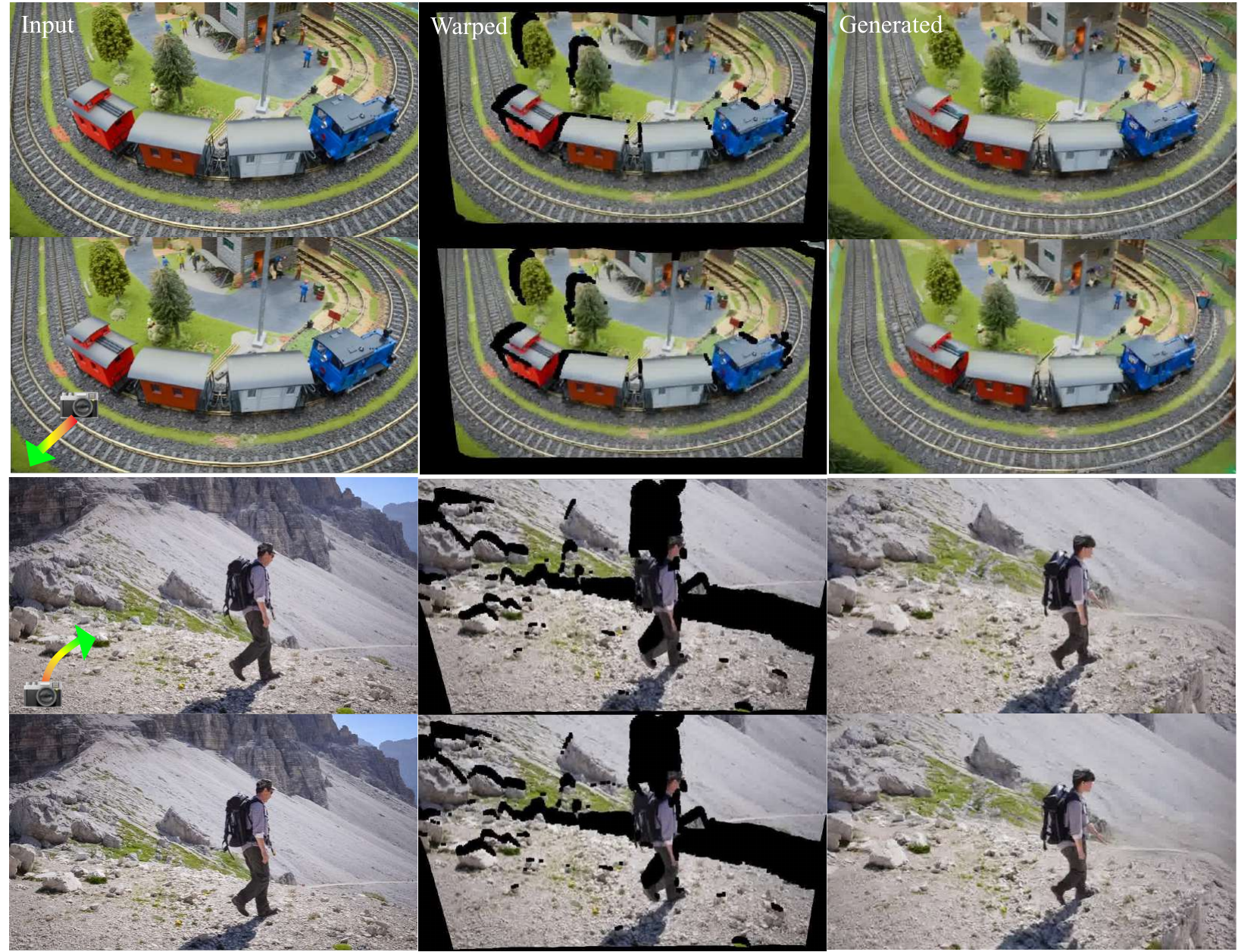}
    \caption{\textbf{Visualization of video camera control process as an inpainting inverse problem on a real video example.} (\textit{"Zoom out", "Translate Up"}.)}
    \label{fig:qualitative_ours}
\end{figure}

\begin{table}[t]
\caption{Quantitative evaluation on the text-guided video inpainting task, showing source video consistency and generation quality.}
\label{tab:inpaint}
\centering
\setlength{\tabcolsep}{3pt}
\resizebox{\columnwidth}{!}{
\begin{tabular}{lccccccc}
\toprule
\textbf{Method} & \textbf{Source Consistency} & \multicolumn{4}{c}{\textbf{Text Alignment}} \\
\cmidrule(r){2-2} \cmidrule(r){3-6}
& \small FVD $\downarrow$ & \small Overall Consistency $\uparrow$ & \small VLM $\uparrow$ & \small CLIP Score $\uparrow$ & \small PickScore $\uparrow$ \\
\cmidrule(r){1-1} \cmidrule(r){2-6}
TrajCrafter~\cite{yu2025trajectorycrafterredirectingcameratrajectory} & 1520.08 & 0.2191 & 0.6140 & 23.88 & 20.24 \\
NVS-Solver~\cite{you2024nvs} & 1648.63 & 0.2240 & \underline{0.6400} & 24.94 & 20.52 \\
Zero4D~\cite{park2025zero4dtrainingfree4dvideo} & 2524.49 &	0.2257 & 0.6260 & 	\textbf{25.83} &	\textbf{20.79} \\
VideoPainter~\cite{bian2025videopainter} & \underline{1432.77} &	\underline{0.2287} & 0.6290 &	\underline{25.26}	& 20.52 \\
\cmidrule(r){1-1} \cmidrule(r){2-6}
\textbf{Ours} (training-free) & \textbf{1422.72} & \textbf{0.2289} & \textbf{0.6510} & 24.98 & \underline{20.56} \\
\bottomrule
\end{tabular}
}
\end{table}
\begin{figure}[!ht]
    \centering
    \includegraphics[width=\linewidth]{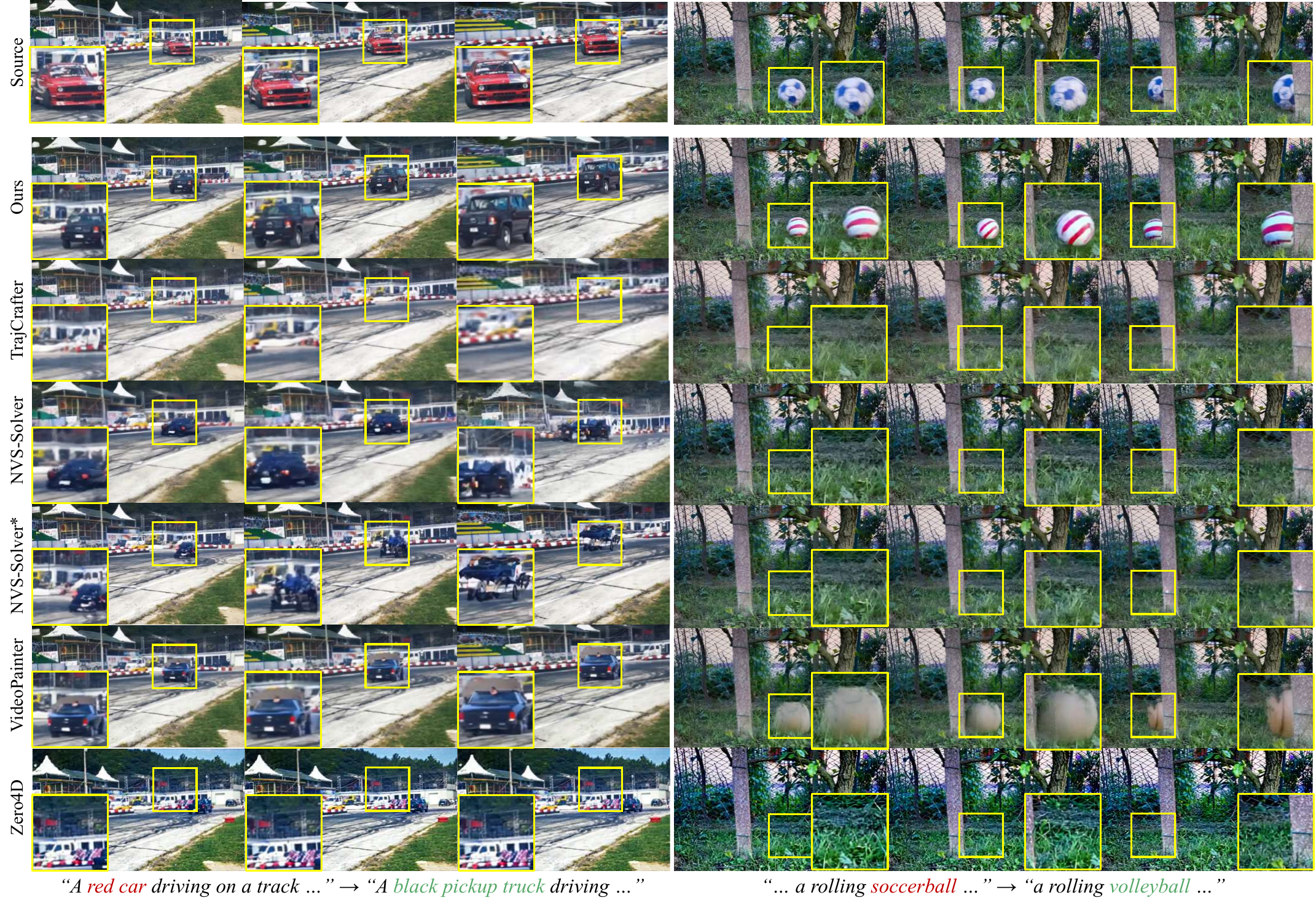}
    \vspace{-0.3cm}
    \caption{\textbf{Qualitative comparison of video inpainting with editing.} InverseCrafter achieves better target text alignment compared to the baselines. NVS-Solver indicates NVS-Solver (post), NVS-Solver* indicates NVS-Solver (dgs).}
    \label{fig:qualitative_inpaint}
\end{figure}

Tab.~\ref{tab:camera-control} reports results on the video camera control task compared to major baselines. Despite requiring neither 4D training data nor VDM fine-tuning, and adding virtually no overhead beyond standard diffusion inference, our method achieves strong performance that outperforms or remains competitive with existing approaches, while providing the best performance when runtime is taken into account (Fig.~\ref{fig:represent}). Furthermore, it delivers consistently strong results across all evaluation metrics, indicating reliable behavior in practical video generation settings. Notably, methods that depend on large-scale VDM training or heavy gradient-based optimization incur high computational cost, whereas our approach attains competitive or superior performance while remaining lightweight.

The qualitative comparisons in Fig.~\ref{fig:qualitative_reangle} further show that baseline training-based or backpropagation methods (TrajectoryCrafter, NVS-Solver (post), TrajectoryAttention, ReCamMaster) generate artifacts (\eg in the trees) or fail to faithfully preserve the source video's content (\eg background car and women's faces). Backpropagation-free methods (NVS-Solver (dgs), Zero4D) fail to capture the video's motion (\textit{e.g.} background car), generate boundary artifacts, or result in highly saturated videos. On the other hand, our method maintains higher fidelity to the source while preserving visual coherence, demonstrating the practical benefit and reliability of our efficient formulation.

To illustrate the generation process on a real video, Fig.~\ref{fig:qualitative_ours} shows how  video camera control is performed by casting the task as an inpainting inverse problem. Fig.~\ref{fig:text_guidance} presents text-conditioned novel content generation results, where the pre-trained VDM prior ensures strong semantic alignment and contextually consistent synthesis under complex geometric manipulation.

\subsubsection{Video Inpainting with Editing}
\label{sec:result_inpaint}
The video inpainting with editing task provides a valid ground-truth latent mask $\vh$ at inference time\footnote{Motivated by this property, a similar training-free projection strategy can also be applied to the video camera control task; see Sec.~\ref{sec:training-free} for details.}, without requiring training of $\maskencoder$. We provide first-frame conditioning input for training-free methods (NVS-Solver, Zero4D) and VideoPainter, as first-frame conditioning is essential for I2V models. The conditioning frame is obtained using an off-the-shelf image inpainting model~\cite{blackforestlabs_flux1filldev_2025}.

The proposed method outperforms the baselines in source distribution and video text alignment metrics in Tab.~\ref{tab:inpaint}. While Zero4D~\cite{park2025zero4dtrainingfree4dvideo} outperforms image based text alignment metrics, this comes from the fact that output video of Zero4D is nearly static as shown in Fig.~\ref{fig:qualitative_inpaint}. In the second example, InverseCrafter uniquely generates the volleyball in the masked region, while also matching the measurement, whereas all baselines fail to generate regarding the target text. In both examples, NVS-Solver (post) deviates a lot from the measurement, NVS-Solver (dgs) and VideoPainter generates boundary artifacts and inferior results. This implies that our method can be used as a general inpainting inverse solver, integrating text descriptions by fully utilizing the pre-trained model's generative prior.
\begin{figure}[!pt]
\centering
\begin{minipage}{0.5\textwidth}
\vspace{-1.3cm}
\centering
\captionof{table}{Quantitative results on VAE reconstruction of different masking strategies.}
\label{tab:ablation_quantitative}
\resizebox{\linewidth}{!}{
\begin{tabular}{lccc}
\toprule
\textbf{Method} & \multicolumn{3}{c}{\textbf{Measurement Consistency}} \\
& \small PSNR $\uparrow$ & \small LPIPS $\downarrow$ & \small SSIM $\uparrow$ \\
\cmidrule(r){1-1} \cmidrule(r){2-4}
{Binary mask resize} & 27.76 & 0.0559 & \textbf{0.8794} \\
Elementwise division & 26.59 & 0.0688 & 0.8458 \\
\cmidrule(r){1-1} \cmidrule(r){2-4}
\textbf{Ours (target)} & \textbf{28.10} & \textbf{0.0526} & 0.8764 \\
\textbf{Ours ($\maskencoder$)} & \underline{27.97} & \underline{0.0515} & \underline{0.8772} \\
\bottomrule
\end{tabular}
}
\end{minipage}
\hfill
\begin{minipage}{0.48\textwidth}
\centering
\includegraphics[width=0.8\linewidth]{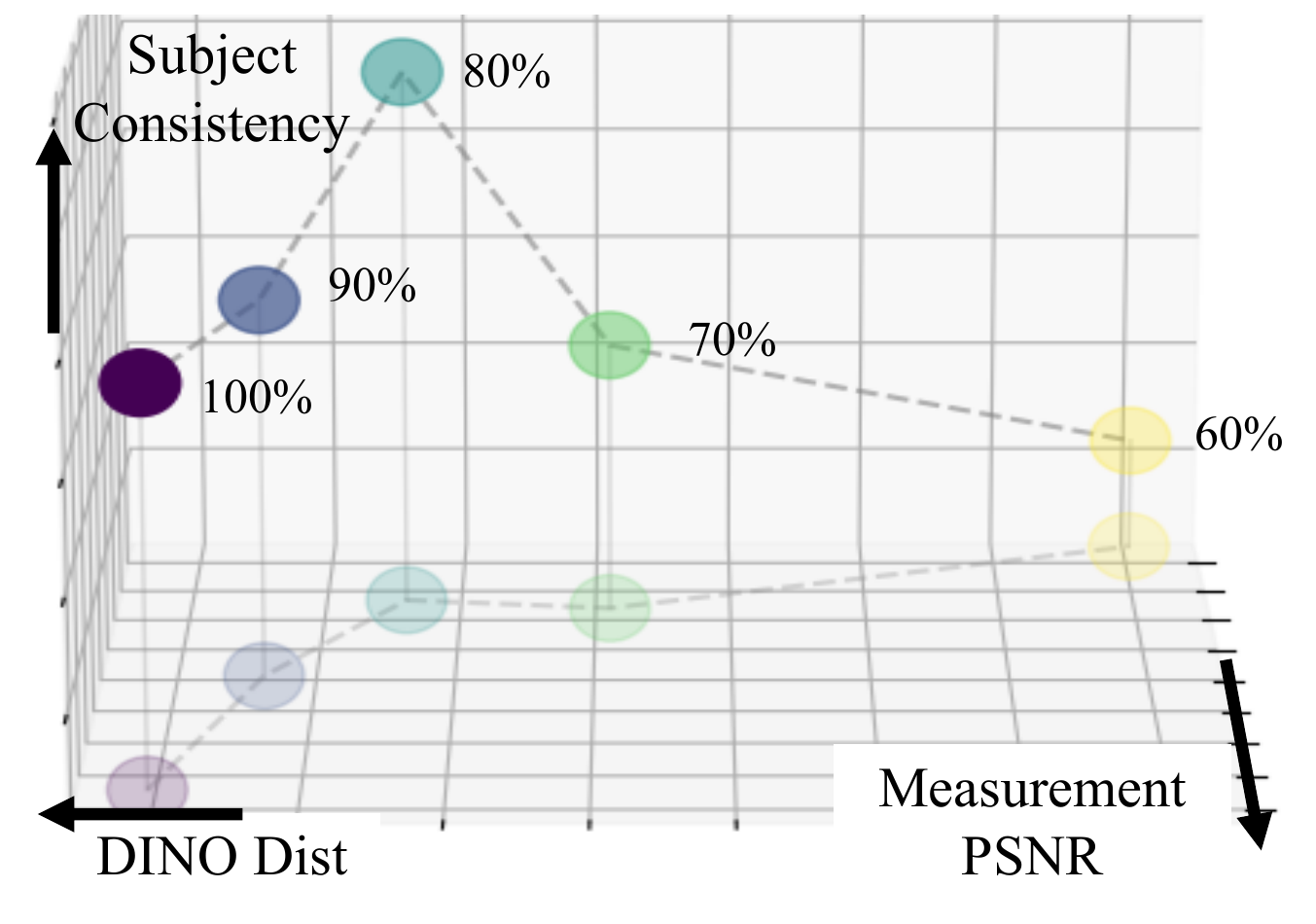}
\vspace{-0.3cm}
\caption{Quantitative comparison on Conjugate Gradient Step Hyperparameter $\alpha$.}
\label{fig:ablation_quantitative}
\end{minipage}
\end{figure}

\begin{figure}[!ht]
    \centering
    \includegraphics[width=0.8\linewidth]{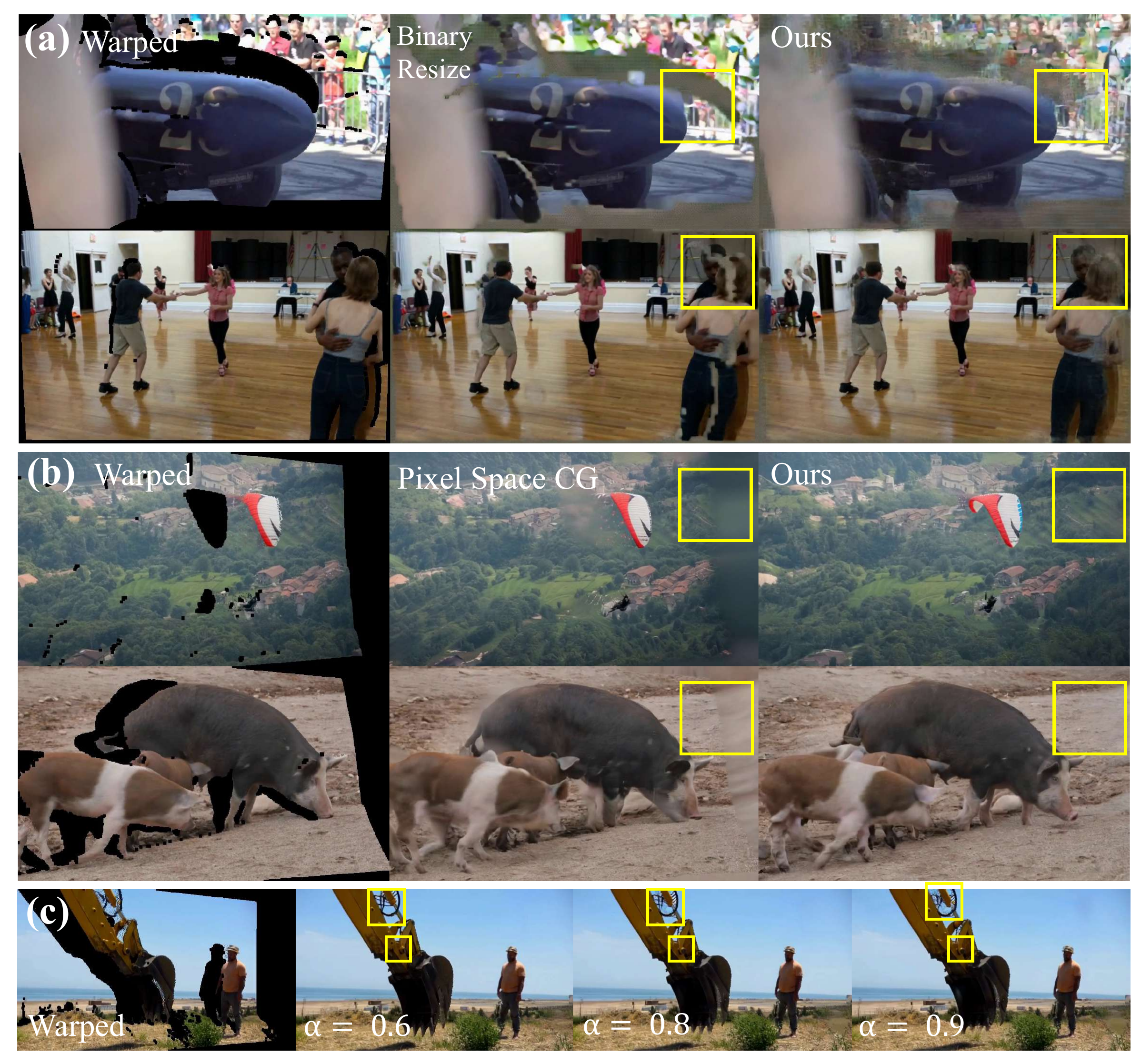}
    \captionof{figure}{\textbf{(a)} VAE reconstruction of binary downsampled mask and ours. \textbf{(b)} Video camera control results of Pixel Space DDS and ours. \textbf{(c)} Ablation results for $\alpha$.}
    \label{fig:ablation}
\end{figure}
\subsection{Ablation Studies}
Tab.~\ref{tab:ablation_quantitative} compares our continuous latent mask formulation with several alternative mask projections on a subset of the DAVIS~\cite{pont20172017} dataset. Conventional binary mask resizes $\times 8$ nearest-neighbor spatially and logical AND downsampling temporally. Binary mask resizing results in a single-channel, conservative mask. Another option is elementwise division, dividing the masked latents by the clean latents. However, this is numerically unstable when the denominator contains small values. 
Overall, our method achieves stronger measurement consistency than these alternatives, indicating that a channel-wise continuous mask better captures how pixel-space occlusions propagate through the VAE encoder.

Additionally, Fig.~\ref{fig:ablation}(b) compares our method with conventional pixel space DDS as in prior methods~\cite{kwon2025visionxlhighdefinitionvideo, kwon2025solvingvideoinverseproblems}, which requires additional VAE encoding and decoding at every timestep, a costly bottleneck that our method entirely avoids. Furthermore, as highlighted by the insets, pixel space Conjugate Gradient introduces severe color and textural inconsistencies. In contrast, our proposed method enables seamless integration with measurement without these artifacts. 

Fig.~\ref{fig:ablation_quantitative} and Fig.~\ref{fig:ablation}(c) illustrate the quantitative and qualitative effect of the conjugate gradient hyperparameter~$\Gamma$ on inpainting performance on UltraVideo~\cite{xue2025ultravideo} dataset, revealing the trade-off between measurement / source consistency and generation fidelity. We set $\Gamma$ as $\Gamma = \{t|t \geq 1 - \alpha\}$. When using a larger $\alpha$ (\textit{i.e.}, applying the measurement consistency step earlier in the diffusion process), the result exhibits stronger consistency with the source measurement and improved content faithfulness. In contrast, a smaller $\alpha$ delays the measurement consistency update to later diffusion steps, emphasizing the model’s generative prior and producing more visually refined outputs. This demonstrates that $\alpha$ effectively controls the balance between the diffusion prior and measurement constraints during posterior sampling, where we observe that setting $\alpha=0.8$ consistently achieves a favorable balance between semantic alignment and structure preservation.
%
%
\vspace{-0.2cm}
\section{Conclusion}
In this paper, we presented a novel and efficient framework for novel-view video generation by casting the task as an inverse problem fully defined in the latent space. The core of our method is to establish operator equivalence between the pixel-space measurement process and its latent-space counterpart via a lightweight latent mask encoder. $\maskencoder$ projects pixel-space masks into continuous, multi-channel latent representations. Our experiments demonstrate that InverseCrafter achieves superior measurement consistency while preserving comparable generation quality in the camera control task. We further show its effectiveness as a general-purpose video inpainting solver for editing tasks. Overall, our framework offers a versatile and lightweight solution for controllable  video synthesis, eliminating the need for computationally expensive VDM fine-tuning and introducing negligible inference-time overhead.

\noindent{\textbf{Limitations and Future Work.}}
Although our solver is efficient, its speed is limited by the multi-step diffusion sampling process. Our method inherits the bias of the pre-trained VDM. The initial warping stage is a major source of artifacts due to its dependence on monocular depth estimates. We expect that performance will further scale with improved depth estimation. The mask encoder is currently tailored to a specific VAE architecture, which may limit applicability to other latent space.
For larger viewpoint changes with more occlusions, the VAE-derived latent-mask target may become less reliable because the masked video contains limited valid visual context. Moreover, the current target mask is constructed by normalizing the latent difference; future work may improve this with more principled latent mask target construction.

\section*{Acknowledgements}
This work was supported by the National Research Foundation of Korea under Grant
RS-2024-00336454, 
Institute of Information \& communications Technology Planning \& Evaluation (IITP) grant funded by the Korea government(MSIT) (No. RS-2022-II220984, Development of Artificial Intelligence Technology for Personalized Plug-and-Play Explanation and Verification of Explanation), 
the Advanced GPU Utilization Support Program funded by the Government of the Republic of Korea (Ministry of Science and ICT) (02-26-01-0404), 
and the AI Computing Infrastructure Enhancement (GPU Rental Support) User Support Program funded by the Ministry of Science and ICT (MSIT), Republic of Korea (RQT-25-120217).

\clearpage  

\bibliographystyle{splncs04}
\bibliography{main}
\clearpage
\onecolumn
\appendix
\section*{\centering \LARGE \textbf{Appendix}}

\section{Related Works}
\subsection{Diffusion and Flow-based Models}
\subsubsection{Diffusion Models}
Diffusion models \cite{ho2020denoising, song2020score} aim to model the data distribution $p(\vx)$, by learning to reverse forward noising process. The forward process, $q(\vx_t | \vx_0)$, gradually adds Gaussian noise to a clean data sample $\vx_0$
\begin{equation} \label{eq:forward_process}
    q(\vx_t | \vx_0) = \mathcal{N}(\vx_t; \alpha_t \vx_0, \sigma_t^2 \mI)
\end{equation}
where $\alpha_t$ and $\sigma_t$ (e.g., $\alpha_t^2 + \sigma_t^2 = 1$) are functions of the noise schedule that control the signal-to-noise ratio at time $t$.
To undo one step of this noising, a neural network $\vepsilon^\theta(\vx_t, t)$ is trained through an epsilon-matching loss:
\begin{align}\label{eq:sm}
    \mathcal{L}_{SM} = \mathbb{E}_{\vepsilon \sim \mathcal{N}(0,\mI), \vx_t \sim p_t} \left\| \vepsilon - \vepsilon^{\theta}(\vx_t, t) \right\|_2^2.
\end{align}
\subsubsection{Flow-based Models}
Continuous Normalizing Flows (CNFs) \cite{papamakarios2021normalizing} model the continuous transformation between two distributions via an ordinary differential equation (ODE). To circumvent the high computational cost and simulation instability of traditional CNF training, Flow Matching (FM) \cite{lipman2023flow} was introduced:
\begin{align}\label{eq:fm}
    \mathcal{L}_{FM} = \mathbb{E}_{t\in [0,1], \vx_t \sim p_t} \| \vv_t(\vx_t) - \vv_t^\theta(\vx_t) \|^2.
\end{align}
The target velocity field $\vv_t(\vx_t)$ in Eq.~\ref{eq:fm} is intractable, as it requires an expectation over the data distribution. Conditional Flow Matching (CFM) \cite{lipman2023flow} solves this by defining a tractable velocity field conditioned on $\vx_0$:
\begin{align}\label{eq:cfm}
    \mathcal{L}_{CFM} = \mathbb{E}_{t\in [0,1], \vx_0 \sim q} \| \vv_t(\vx_t|\vx_0) - \vv_t^\theta(\vx_t) \|^2.
\end{align}
With a velocity network $\vv_t^\theta$ trained on this objective, generation is performed by solving the ODE $d\vx_t = \vv_t^\theta(\vx_t) dt$ from $t=1$ to $t=0$. A prominent variant is Rectified Flow \cite{liu2023flow} which employs a linear interpolation path ($a_t=t$ and $b_t=1-t$), resulting in $\vv_t(\vx_t|\vx_0)= \vx_1 - \vx_0$. Crucially, when the prior distribution $p_1$ is Gaussian, FM is equivalent to diffusion models\footnote{Set $\alpha_t = 1 - t, \sigma_t = t$ in \eqref{eq:forward_process}.}. We therefore use these terms interchangeably henceforth.
\subsubsection{Latent Diffusion models}
Training and inference of diffusion models are computationally expensive on high-resolution data. To improve efficiency, these processes can be reformulated in a compressed latent space, as popularized by Latent Diffusion Models (LDMs) \cite{rombach2022high} and used in most modern diffusion models~\cite{videoworldsimulators2024,flux,wan2025wanopenadvancedlargescale}. A pre-trained Variational Autoencoder (VAE) is used to map data from the pixel space to the latent space ($\vz = \mathcal{E}(\vx)$) and back ($\hat{\vx} = \mathcal{D}(\vz)$). The generative model ($\vepsilon_\theta$ or $\vv_t^\theta$) is then trained entirely on the latent representations $\vz$.

\subsection{Diffusion Inverse Solvers}
Given a pre-trained generative model as a prior $p(\vx)$, Diffusion Inverse Solvers (DIS) aim to solve inverse problems in Eq.~\ref{eq:inverse}. This is achieved by guiding the reverse sampling process to sample from the posterior distribution $p(\vx|\vy)$. A significant advantage of this paradigm is its zero-shot generalization capability; the same pre-trained model $p(\vx)$ can be applied to various forward operators $\mathcal{A}$ at inference time without re-training. Broadly, these solvers can be categorized by their guidance mechanism.
\subsubsection{Backpropagation-Based Solvers}
One prominent line of work, including Diffusion Posterior Sampling (DPS) \cite{chung2023diffusion}, guides the sampling at each step $t$ by using the gradient of the data consistency term, $\nabla_{\vx_t} \log p(\vy|\vx_t)$. This is often approximated by $\nabla_{\vx_t} \|\vy - \mathcal{A}(\hat{\vx}_{0|t})\|^2$, where $\hat{\vx}_{0|t}$ is the predicted clean data from $\vx_t$. While effective, these methods are computationally prohibitive as they require expensive backpropagation through the deep neural network $\vepsilon^\theta$ (or $\vv^\theta$) at every sampling step.

\subsubsection{Backpropagation-Free Solvers}
To address the high computational cost, a second line of work seeks to avoid this backpropagation. Decomposed Diffusion Sampling (DDS) \cite{chung2024decomposed}, for example, synergistically combines diffusion sampling with classical optimization methods. For linear inverse problems, this update is equivalent to solving a proximal optimization problem:
\begin{align}
\label{eq:prox}
    \min_\vx \frac{\gamma}{2}\|\vy - \mathcal{A}(\vx)\|_2^2 + \frac{1}{2}\|\vx - \hat{\vx}_{0|t}\|_2^2,
\end{align}
which can be solved efficiently using a small number of Conjugate Gradient (CG) steps. This approach eliminates the need to compute the costly Manifold-Constrained Gradient (MCG), resulting in dramatic accelerations.

\subsection{Novel View Synthesis}
Early efforts to imbue VDMs with precise camera control involved modifying the model architecture or training process. This included training external modules to inject conditioning information such as camera parameters \cite{he2024cameractrl, li2025realcam}, conditioning the model on explicit geometric representations like Pl\"{u}cker embeddings \cite{wu2024cat4d, bahmani2024vd3d}, or inflating the attention mechanisms of the base VDM \cite{xu2024camco, watson2024controlling, sun2024dimensionx, bai2024syncammaster}.

\subsubsection{Target-View Video Recapture}
More recently, the problem has been effectively framed as target-view recapture task given an input video. The first and most common direction involves fine-tuning a pre-trained VDM. This includes methods that train the model to directly accept camera parameters as input conditions \cite{van2024generative, wu2024cat4d, bai2025recammastercameracontrolledgenerativerendering}, or methods that tries to solve a 3D-aware inpainting problem \cite{gu2025diffusionshader3dawarevideo, xiao2024trajectory, jeong2025reangleavideo4dvideogeneration}. These methods warp the source video according to the target trajectory and then fine-tune the VDM to realistically inpaint the resulting occluded or disoccluded regions. 
The second, training-free methods adapt pre-trained VDMs as powerful generative priors, typically by reformulating the task as an inverse problem that can be solved at inference time \cite{you2024nvs, park2025zero4dtrainingfree4dvideo}. Our work builds upon this paradigm by developing a more efficient and robust latent-space solver for this task.

\section{Implementation Details}
\subsection{Text-guided Inpainting Dataset}
The Text-guided Inpainting dataset consists of video clips, target masks, source prompts, and target prompts. We selected 20 videos and target mask pairs from the DAVIS~\cite{Caelles_arXiv_2019} dataset. In each pair, the foreground in the mask is treated as the inpainting target. For source prompt generation, we used the BLIP-2~\cite{li2023blip} model to generate a caption describing the middle frame of each video. The target prompts were then generated using the \texttt{gpt-5-mini-2025-08-07}~\cite{openai_gpt5mini_doc_2025} model, which took the source prompt as input along with the following instruction prompt (Fig.~\ref{fig:inpainting-dataset}). A total of five target prompts were generated for each video, resulting in a dataset containing 125 samples.

\begin{figure}[!ht]
  \centering
  \small
  \begin{tcolorbox}[
    colback=gray!5!white,
    colframe=black,
    title=\textbf{Prompt for Inpainting Target Prompt Generation}
  ]

You are an AI assistant for generating paired text prompts for real image editing tasks. Your goal is to modify a given text description by replacing an object with other while strictly following these rules:
\\
\\
1) Modify the primary object that appears in the original prompt. There is only one primary object in the image.\\
2) The replacement must be significantly different from the original concept but contextually appropriate. Avoid unrealistic substitutions (e.g., changing `rabbit on grass' to `rocket on grass').\\
3) Ensure diversity in word choices across different modifications.\\
4) Preserve all other words exactly as they are. Do not change sentence structure, introduce new elements, or modify additional details.\\
5) Do not provide any additional words—output only the modified text description.\\
6) Editing is done by masking the primary object and filling in only that area (text-guided inpainting). In other words, it is not possible if the silhouette of the target object differs too much, so please avoid creating target prompts that result in such discrepancies.\\
\\
The output format must be like this: ["<modified prompt1>", "<modified prompt2>", "<modified prompt3>", "<modified prompt4>", "<modified prompt5>"]. You must generate **5** different versions of the modified prompt. 
\\

original prompt: \texttt{\{text\}}

  \end{tcolorbox}
  \caption{Prompt used for text-guided inpainting prompt generation with \texttt{GPT-5-mini}. Given a source prompt describing the initial video frame, the model is instructed to generate diverse and contextually consistent target prompts to construct the inpainting dataset.}
  \label{fig:inpainting-dataset}
\end{figure}

\subsection{Evaluation Metric}
\label{sec:vllm}
For the quantitative comparison, we evaluate following metrics following the evaluation code \footnote{\url{https://github.com/cure-lab/PnPInversion/tree/main/evaluation}} \footnote{\url{https://github.com/yuvalkirstain/PickScore}} \footnote{\url{https://github.com/JunyaoHu/common_metrics_on_video_quality}} \footnote{\url{https://github.com/Vchitect/VBench}}:
\begin{enumerate}
    \item Measurement PSNR : We compute the PSNR by excluding the masked region, resulting in the measurement PSNR.
    \item Measurement LPIPS : We measure the LPIPS~\cite{blau2018perception}, which is defined as distance between feature maps of pre-trained VGG network, by excluding the masked region.
    \item Measurement SSIM : We compute the structural similarity~\cite{wang2004image} by excluding the masked region.
    \item DINO distance : We evaluate source distribution consistency by computing the average cosine distance between DINO-ViT~\cite{caron2021emerging} features extracted from the generated video and the source video.
    \item FVD : We report the Fréchet Video Distance (FVD)~\cite{unterthiner2018towards} to assess the overall quality and temporal coherence of the generated videos, using features from a pre-trained I3D network.
    \item VBench: We calculate subject consistency, temporal flickering, motion smoothness for Quality Score and overall consistency for Semantic Score, calculate the Total Score based on the metric weights $\&$ normalization method in VBench.
    \item CLIP-score : We report the similarity between features embedded by pre-trained CLIP~\cite{radford2021learning}\footnote{We use CLIP ViT-base-patch16.} image encoder and text encoder.
    \item Pick-score : To measure the alignment between the generated video and the target text prompt, we use the Pick-score~\cite{kirstain2023pickapicopendatasetuser} to predict human preferences, providing a proxy for both text-video alignment and aesthetic quality.
    \item VLM : We discuss about this in the following section.

\end{enumerate}

\subsubsection{VLM metric}
To assess both the visual quality of the generated videos and their alignment with the target text prompt, we employ a Vision Language Model (VLM)–based evaluation metric. Specifically, we use \texttt{Qwen2.5-VL-3B-Instruct}~\cite{bai2025qwen2}, which is capable of processing video input natively. The VLM is prompted to evaluate each video–prompt pair across three criteria in equal weight: (1) faithfulness to the prompt, (2) visual quality, and (3) temporal coherence. The model outputs a score between 0 and 10, which we normalize to the range of 0 to 1 for reporting. The full evaluation prompt provided to the VLM is shown in Fig.~\ref{fig:vllm-metric}.

\begin{figure}[!ht]
  \centering
  \small
  \begin{tcolorbox}[
    colback=gray!5!white,
    colframe=black,
    title=\textbf{Prompt for VLM Evaluation Metric}
  ]

You are an expert video evaluator. Your task is to judge a video based on three equally weighted aspects: \\

1. **Faithfulness to Prompt**: Does the video accurately reflect the user’s input prompt in terms of objects, attributes, style, and composition?\\
2. **Visual Quality**: Is the video clear, sharp, and free from any unrealistic artifacts or distortions?\\
3. **Temporal Coherence**: Is the video smooth and coherent over time, without abrupt changes or jarring transitions?\\

Please rate this video on a scale of 0-10 (10 being perfect) and explain your reasoning. Please put your score in <score> score </score>. Prompt: \texttt{\{pr\}}

  \end{tcolorbox}
  \caption{Evaluation prompt used for VLM-based scoring with Qwen2.5-VL-3B-Instruct. The model is given a generated video and its corresponding text prompt, and returns a score reflecting prompt faithfulness, visual quality, and temporal coherence.}
  \label{fig:vllm-metric}
\end{figure}

\section{Further discussion on comparison with SILO}

SILO~\cite{raphaeli2025silo} learns a general nonlinear degrader \(H_\theta(\hat{\vz}_{0|t},t)\) that maps a predicted clean latent to the encoded measurement \(\vw=\mathcal{E}(\vy)\). Thus, SILO remains a nonlinear inverse problem and requires backpropagating through \(H_\theta\) at every sampling step. In contrast, our inpainting setting is structured as \(\vy=\vm\odot\vx\). We exploit this structure by learning only the parameters of the latent operator: a continuous channel-wise latent mask $\vh = \maskencoder(\vy)$. This enables (i) plug-and-play use of standard linear inverse solvers, (ii) avoids backpropagation through a degradation network or the VDM, and (iii) requires evaluating $\maskencoder$ only once.

Furthermore, direct extension of SILO is not straightforward in our setting. SILO trains \(H_\theta\) by applying a known degradation to clean samples and matching the resulting encoded measurement. In video recapture, clean target-view videos are unavailable. In contrast, our double-reprojection strategy provides supervision for the latent mask operator. Second, SILO is only evaluated on fixed degradations (\textit{e.g.} center-box masks), whereas our task involves sample-dependent, temporally varying warping masks. Extension of SILO on random masks gets naively resized latent mask as an input condition, reintroducing the 3D-VAE mismatch that our method is designed to address (L188-199).  In contrast, our $\vh$ is trained from the actual VAE-induced corruption: the target mask is computed from the discrepancy between the clean latent \(\mathcal{E}(\vx)\) and the corrupted latent \(\mathcal{E}(\vm\odot\vx)\). Thus, the nonlinear effect of encoding a masked video through the 3D VAE is directly used as supervision.
Tab.~\ref{tab:ablation_quantitative} further supports that the proposed method better captures how pixel-space occlusions propagate through the VAE encoder than prior naive binary sub-sampling.
Finally, contrary to SILO,  our method even supports a training-free variant, where masks are computed at run time with minimal performance degradataion  (Sec.~\ref{sec:training-free}).

\section{Training-free Variant of InverseCrafter}
\label{sec:training-free}
While the learned $\maskencoder$ remains our primary method, our framework also supports a training-free variant for latent mask generation. For text-guided video inpainting, we directly compute the latent mask from the input video and its masked counterpart, since the input mask directly corresponds to the video being edited.

Camera control creates a harder case. Because we seek to generate the target novel-view video itself, we do not have target-view ground truth and cannot construct a perfectly matched latent mask from the target view. At inference time, we only have the target warping mask $\vm_{\text{warp}}$ and the source-view video $\vx^{src}$.

If we apply the target warping mask to the clean source video, rather than to the warped video, and compute the proxy latent mask as
\begin{equation}
    \vh = \mathrm{f}\!\left(\mathcal{E}(\vx^{src}) - \mathcal{E}(\vm_{\text{warp}} \odot \vx^{src})\right).
\end{equation}
However, the mask $\vm_{\text{warp}}$ and the source video $\vx^{src}$ do not share the same camera viewpoints, so $\vm_{\text{warp}} \odot \vx^{src}$ does not form a physically valid observation of either the source or the target view. Even so, the proxy works for two reasons. First, the latent mask mainly needs to encode where content stays visible and where content disappears, and the warping mask preserves that spatial localization. Second, background semantics usually change only moderately under the camera motions that we consider, so the VAE can still turn the masked source video into a useful channel-wise soft mask even when local semantics do not perfectly match the target view.
We therefore treat the training-free variant as a practical fallback, while we keep the learned $\maskencoder$ as our default choice because it models the latent masking process more faithfully. A key advantage of InverseCrafter lies in this flexibility: the same framework supports both learned and training-free latent mask generation.

Tab.~\ref{tab:novel_traj} shows that the training-free variant remains highly competitive. Compared with the learned $\maskencoder$, the training-free variant incurs only a marginal drop in measurement consistency and even slightly improves some source-distribution metrics. Given that it removes all training cost, we view this gap as minor in practice.

\section{Additional Results}
\subsection{InverseCrafter + different Diffusion Inverse Solvers}
\label{sec:solvers}
\begin{figure}[!pt]
    \centering
    \includegraphics[width=\linewidth]{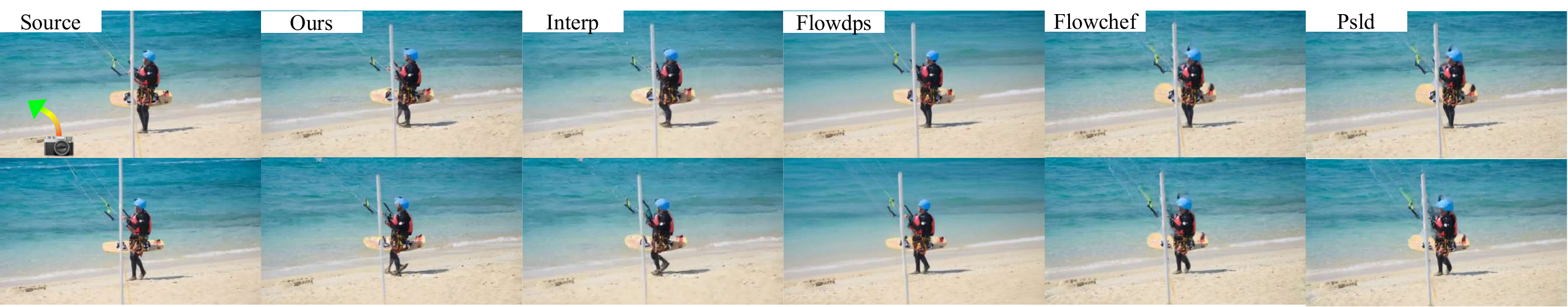}
    \vspace{-0.6cm}
    \caption{InverseCrafter is solver-agnostic and enables the use of \textit{any} diffusion inverse solver directly in the latent space.}
    \vspace{-0.3cm}
    \label{fig:solvers}
\end{figure}

Our work focuses on making the inpainting step highly efficient, whereas existing methods rely on costly training and inference-time backpropagation. Also, common latent inverse solvers that utilize DDS require VAE decoding at every sampling step. InverseCrafter reformulates the nonlinear measurement induced by the decoder into a linear form in latent space via a latent mask, enabling versatility with \textit{any} inverse solvers~\cite{kim2025flowdps, patel2024steeringrectifiedflowmodels, rout2023solvinglinearinverseproblems}. This reformulation creates a setting in which linear-only DDS can be applied with minimal overhead in latent models by avoiding model backpropagation or repeated evaluation of the VAE. This enables inpainting with negligible overhead, which is critical in practice considering the scale and computational demands of contemporary VDMs. We demonstrate compatibility across different solvers in Fig.\ref{fig:solvers}. Tab.~\ref{tab:novel_traj} further shows quantitative results of InverseCrafter + FlowDPS~\cite{kim2025flowdps}.

\subsection{InverseCrafter + different backbone VDM}
Tab.~\ref{tab:novel_traj} shows compatibility of InverseCrafter with a different backbone model~\cite{yang2024cogvideox}. The mask encoder architecture is currently tailored to the VDM's VAE architecture. Nevertheless, our framework remains compatible with different VDM backbone.

\subsection{Visualization of latent mask}
\begin{figure}
    \centering
    \includegraphics[width=0.8\linewidth]{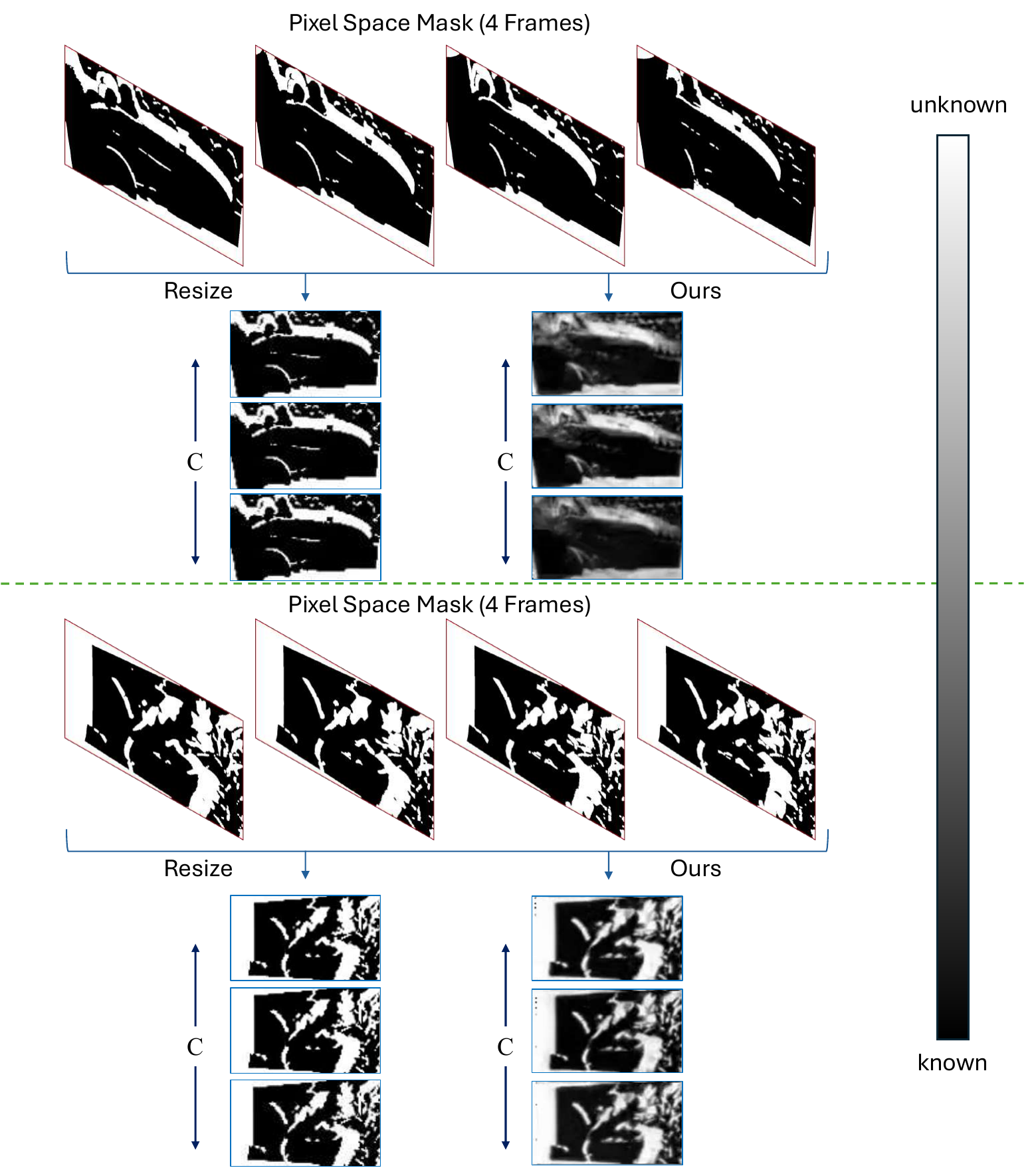}
    \caption{\textbf{Comparison of latent mask generation process.}}
    \label{fig:latent_mask}
\end{figure}

Fig.~\ref{fig:latent_mask} compares the latent masks produced by conventional binary downsampling against those from InverseCrafter. The naive resizing approach yields a single-channel, binary mask that is uniformly broadcast across all $C$ channels. In contrast, our method generates a continuous, $C$-channel mask, preserving distinct spatio-temporal information for each latent feature.

\subsection{Runtime Comparison}
In this section, we validate the computational efficiency of InverseCrafter by comparing its wall-clock inference time against the baseline Video Diffusion Model (VDM). Consistent with the claims made in the main text, our results demonstrate that the proposed method incurs negligible runtime overhead. Our evaluation focuses specifically on the execution time of the diffusion transformer during the latent-space ODE solving process. Experiments were conducted on a workstation equipped with an AMD EPYC 7543 32-Core Processor and an NVIDIA RTX A6000 GPU.
\begin{table}[h!]
    \centering
    \caption{Runtime Comparison.}    
    \begin{tabular}{|l|c|c|}
        \hline
        \textbf{Method} & \textbf{Ours ($\alpha=0.8$)} & \textbf{Base ($\alpha=0.0$)} \\
        \hline
        Runtime (sec) & 71 & 70 \\
        \hline
    \end{tabular}
    \label{tab:runtime_comparison}
\end{table}

\subsection{Novel Trajectory Evaluation}
Tab.~\ref{tab:novel_traj} evaluates generalization to unseen camera trajectories on a subset of the DAVIS dataset. InverseCrafter generalizes to unseen trajectories, since the learned latent mask models the visibility rather than specific camera motions.

\begin{table}[h]
\centering
\caption{Novel trajectory evaluation and backbone/solver ablation.}
\resizebox{\linewidth}{!}{
\begin{tabular}{lcccc|cccc}
\toprule
 & TC & NVS-Solver & CogNVS & GCD & Ours & Ours & Ours & Ours \\
 &  &  &  &  &  & (training-free) & (CogVideoX) & (FlowDPS) \\
\midrule

PSNR
& 25.44
& 21.38
& 14.62
& 15.60
& 25.60
& 25.48
& 26.46
& \textbf{27.43} \\

LPIPS
& 0.087
& 0.153
& 0.442
& 0.312
& 0.077
& 0.078
& 0.095
& \textbf{0.070} \\

DINO
& 0.036
& 0.037
& 0.101
& 0.068
& \textbf{0.033}
& \textbf{0.033}
& 0.035
& 0.035 \\

VBench
& 0.856
& 0.857
& 0.794
& 0.840
& 0.851
& 0.852
& \textbf{0.862}
& 0.853 \\

\bottomrule
\end{tabular}
}
\label{tab:novel_traj}
\end{table}

\subsection{Additional Qualitative Comparison}
\begin{figure}
    \centering
    \includegraphics[width=0.8\linewidth]{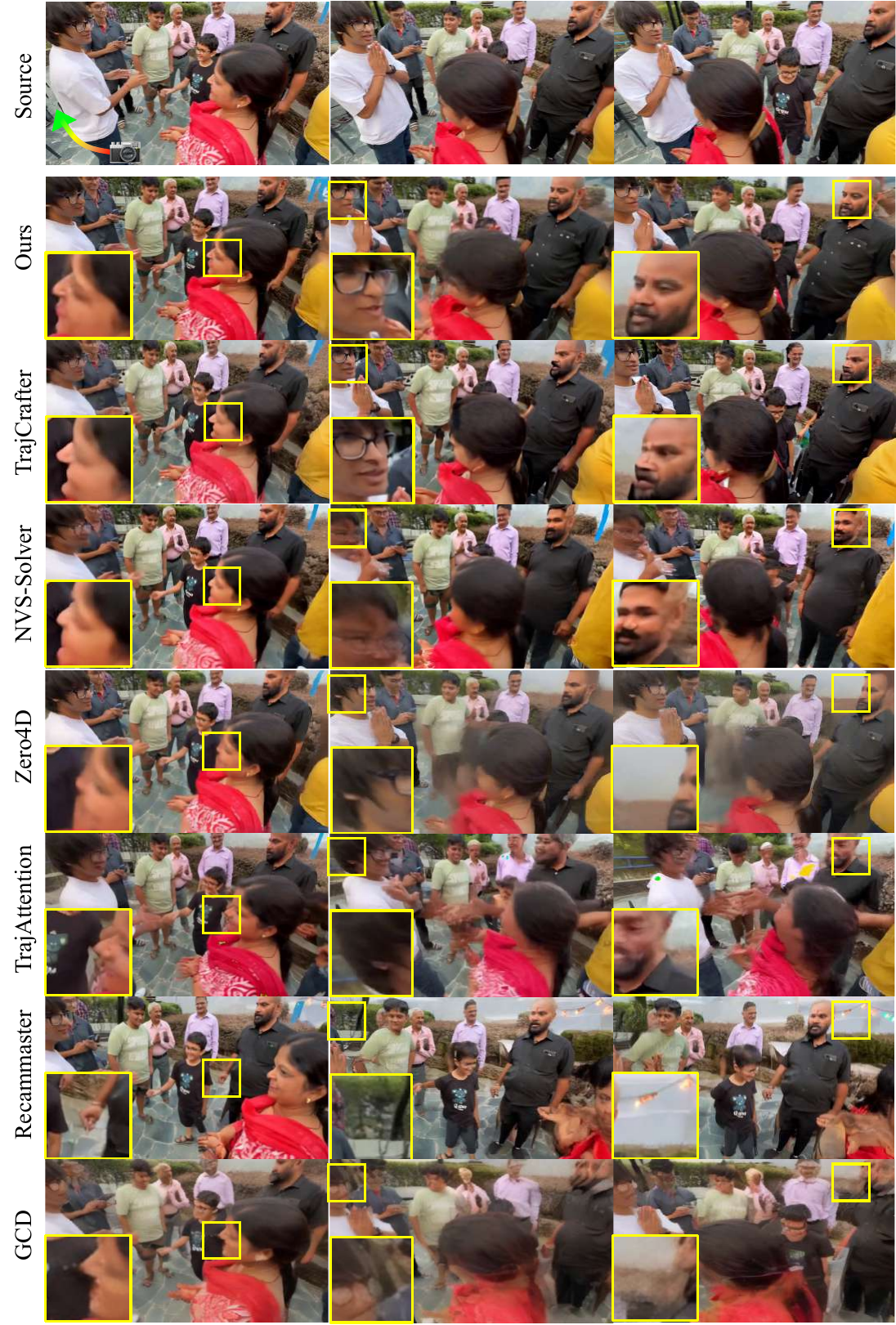}
    \caption{\textbf{Additional qualitative comparison of video camera control.}}
    \label{fig:supple_qualitative_reangle1}
\end{figure}

\begin{figure}
    \centering
    \includegraphics[width=0.8\linewidth]{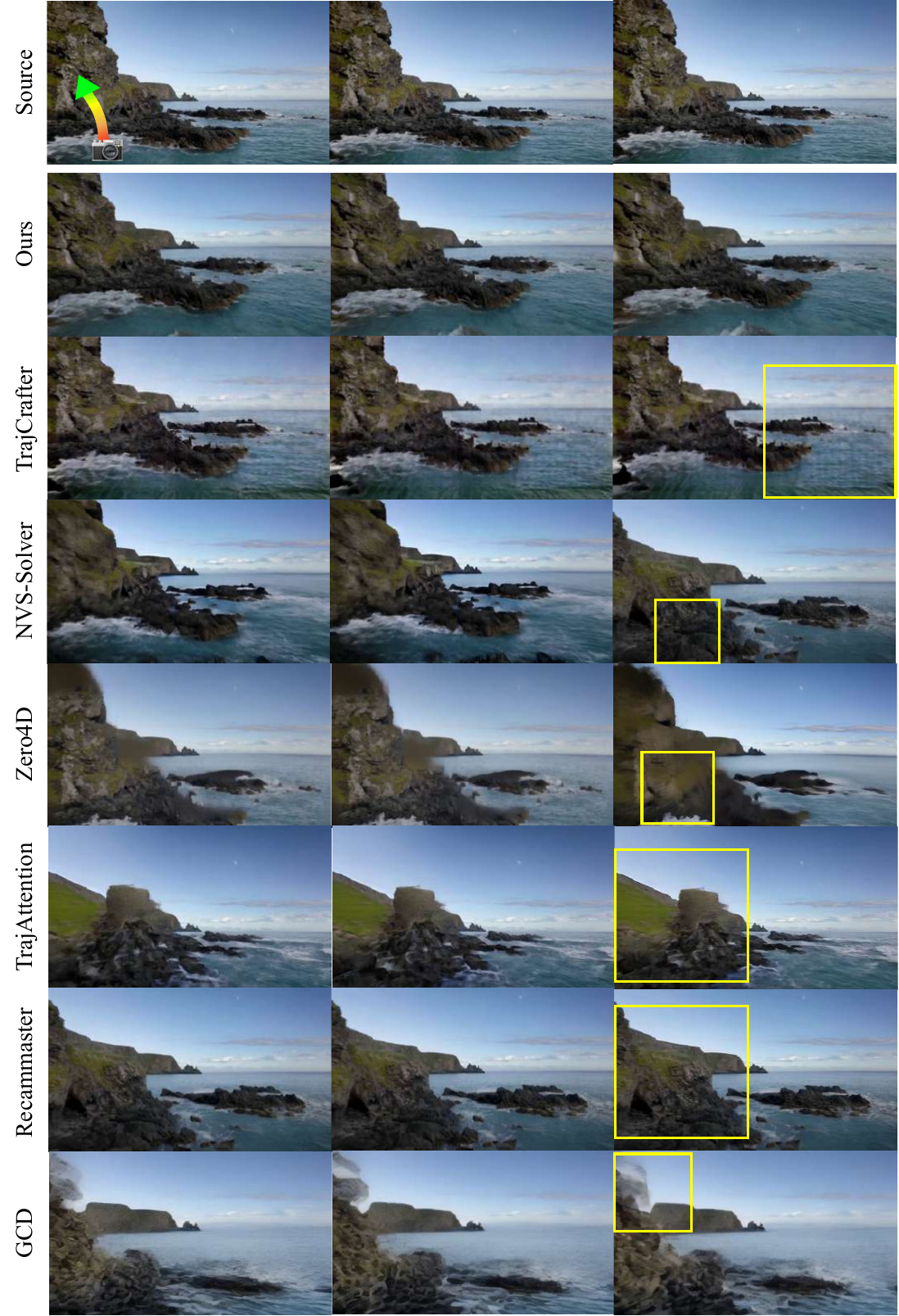}
    \caption{\textbf{Additional qualitative comparison of video camera control.}}
    \label{fig:supple_qualitative_reangle2}
\end{figure}

\begin{figure}
    \centering
    \includegraphics[width=0.9\linewidth]{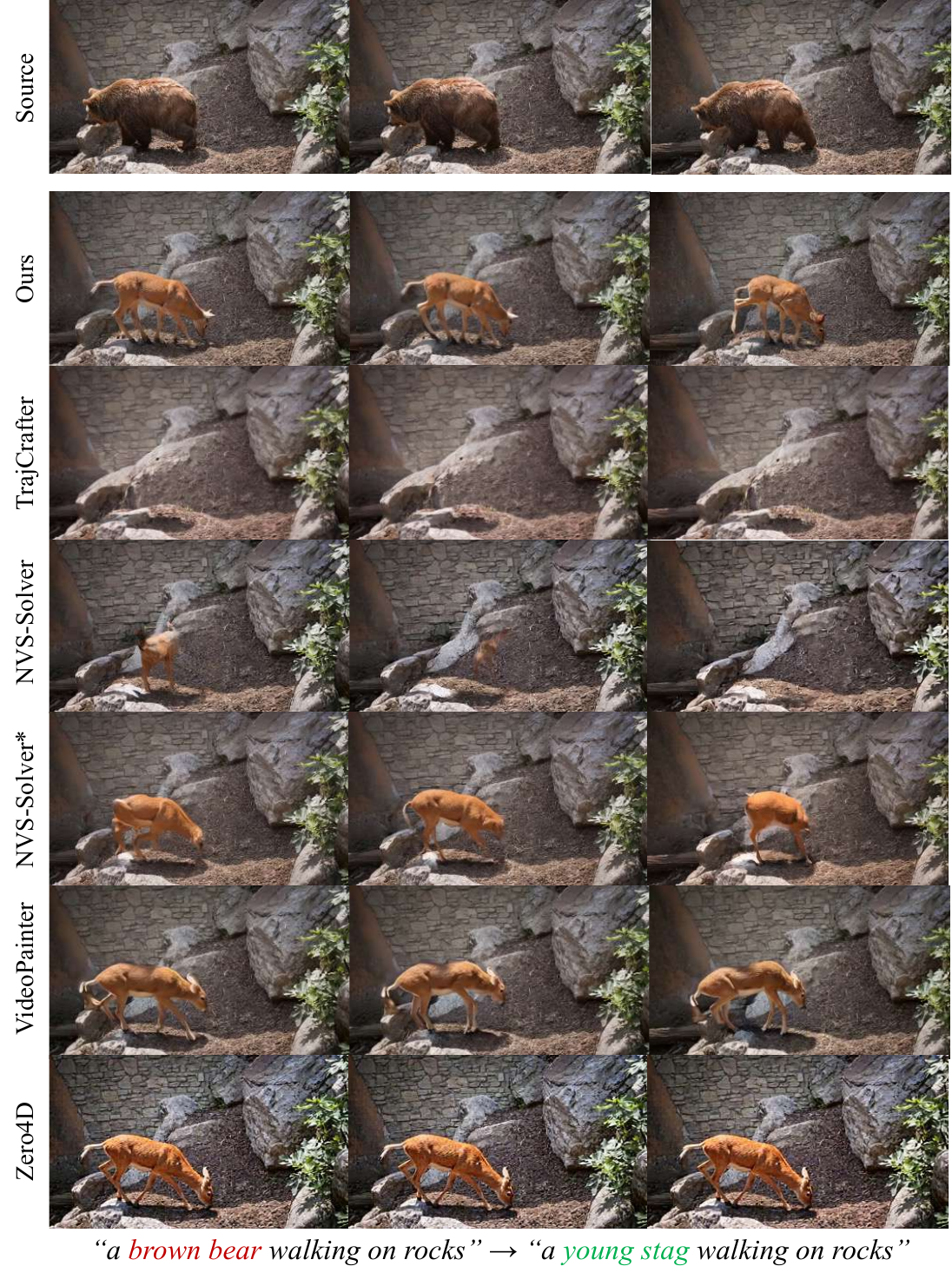}
    \caption{\textbf{Additional qualitative comparison of video inpainting with editing.} NVS-Solver indicates NVS-Solver (post), NVS-Solver* indicates NVS-Solver (dgs).}
    \label{fig:supple_qualitative_inpaint1}
\end{figure}

\begin{figure}
    \centering
    \includegraphics[width=0.9\linewidth]{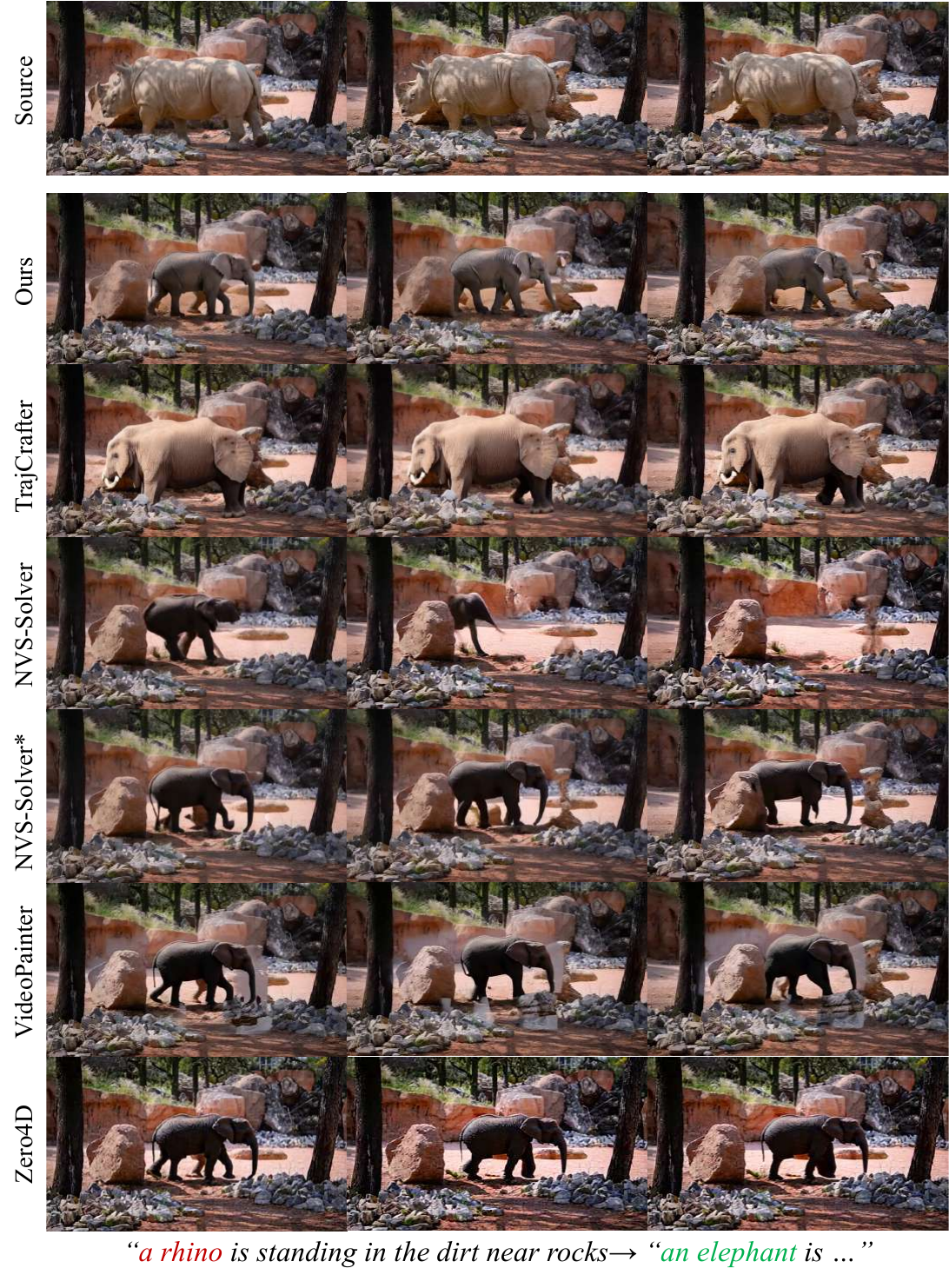}
    \caption{\textbf{Additional qualitative comparison of video inpainting with editing.} NVS-Solver indicates NVS-Solver (post), NVS-Solver* indicates NVS-Solver (dgs).}
    \label{fig:supple_qualitative_inpaint2}
\end{figure}

We provide further qualitative results, demonstrating our method's performance against baseline methods. Fig.~\ref{fig:supple_qualitative_reangle1} and \ref{fig:supple_qualitative_reangle2} present results for the video camera control task, fig.~\ref{fig:supple_qualitative_inpaint1} and \ref{fig:supple_qualitative_inpaint2} shows results for the video inpainting with editing task.

\label{sec:mask_viz}

\end{document}